\documentclass[sigconf]{acmart}
\usepackage{graphicx}
\usepackage{multirow}
\usepackage{enumitem}
\usepackage{subfigure}
\usepackage{microtype}
\usepackage{float}

\AtBeginDocument{%
  \providecommand\BibTeX{{%
    \normalfont B\kern-0.5em{\scshape i\kern-0.25em b}\kern-0.8em\TeX}}}

\copyrightyear{2024}
\acmYear{2024}
\setcopyright{acmlicensed}\acmConference[MM '24]{Proceedings of the 32nd ACM International Conference on Multimedia}{October 28-November 1, 2024}{Melbourne, VIC, Australia}
\acmBooktitle{Proceedings of the 32nd ACM International Conference on Multimedia (MM '24), October 28-November 1, 2024, Melbourne, VIC, Australia}
\acmDOI{10.1145/3664647.3680962}
\acmISBN{979-8-4007-0686-8/24/10}

\begin{document}

\title{Diffusion Domain Teacher: Diffusion Guided Domain Adaptive Object Detector}

\settopmatter{authorsperrow=4}

\author{Boyong He}
\authornote{Contribute equally to the work.}
\affiliation{%
  \institution{Xiamen University}
  \department{Institute of Artifcial Intelligence}
  % \streetaddress{1 Th{\o}rv{\"a}ld Circle}
  \city{Xiamen}
  \country{China}}
\email{boyonghe@stu.xmu.edu.cn}

\author{Yuxiang Ji}
\authornotemark[1]
\affiliation{%
  \institution{Xiamen University}
  \department{Institute of Artifcial Intelligence}
  % \streetaddress{1 Th{\o}rv{\"a}ld Circle}
  \city{Xiamen}
  \country{China}}
\email{yuxiangji@stu.xmu.edu.cn}

\author{Zhuoyue Tan}
\affiliation{%
  \institution{Xiamen University}
  \department{Institute of Artifcial Intelligence}
  % \streetaddress{1 Th{\o}rv{\"a}ld Circle}
  \city{Xiamen}
  \country{China}}
\email{tanzhuoyue@stu.xmu.edu.cn}

\author{Liaoni Wu}
\authornote{Corresponding author.}
\affiliation{%
  \institution{Xiamen University}
  \department{Institute of Artifcial Intelligence}
  \department{School of Aerospace Engineering}
  % \streetaddress{1 Th{\o}rv{\"a}ld Circle}
  \city{Xiamen}
  \country{China}}
\email{wuliaoni@xmu.edu.cn}
\renewcommand{\shortauthors}{Boyong He, Yuxiang Ji, Zhuoyue Tan, and Liaoni Wu.}

\begin{abstract}

  Object detectors often suffer a decrease in performance due to the large domain gap between the training data (source domain) and real-world data (target domain). Diffusion-based generative models have shown remarkable abilities in generating high-quality and diverse images, suggesting their potential for extracting valuable feature from various domains. To effectively leverage the cross-domain feature representation of diffusion models, in this paper, we train a detector with frozen-weight diffusion model on the source domain, then employ it as a teacher model to generate pseudo labels on the unlabeled target domain, which are used to guide the supervised learning of the student model on the target domain. We refer to this approach as \textit{Diffusion Domain Teacher} (\textbf{DDT}). By employing this straightforward yet potent framework, we significantly improve cross-domain object detection performance without compromising the inference speed. Our method achieves an average mAP improvement of 21.2\% compared to the baseline on 6 datasets from three common cross-domain detection benchmarks (\textit{Cross-Camera, Syn2Real, Real2Artistic)}, surpassing the current state-of-the-art (SOTA) methods by an average of 5.7\% mAP. Furthermore, extensive experiments demonstrate that our method consistently brings improvements even in more powerful and complex models,  highlighting broadly applicable and effective domain adaptation capability of our DDT. The code is available at \textit{\color{blue}\href{https://github.com/heboyong/Diffusion-Domain-Teacher}{https://github.com/heboyong/Diffusion-Domain-Teacher}}.
\end{abstract}

%   <ccs2012>
%   <concept>
%   <concept_id>10010147.10010178.10010224.10010240.10010241</concept_id>
%   <concept_desc>Computing methodologies~Image representations</concept_desc>
%   </concept>
%   <concept>
%   <concept_id>10010147.10010178.10010224.10010245.10010250</concept_id>
%   <concept_desc>Computing methodologies~Object detection</concept_desc>
%   </concept>
%   </ccs2012>
  
% \ccsdesc[500]{Computing methodologies~Image representations}
% \ccsdesc[500]{Computing methodologies~Object detection}




\maketitle

\section{Introduction}

% check
%%%%%%%%%%%%%%%%%%%%%%%%%%%%%%%%%%%%%%%%%%%%%%%%%%%%%%%%%%%%%%%%%%%%%%%%%%%%%%%%%%%%%%%%%%%%%%%
\begin{figure}[t]
   % 调整间距
  \centering
  \includegraphics[width=1.0\linewidth]{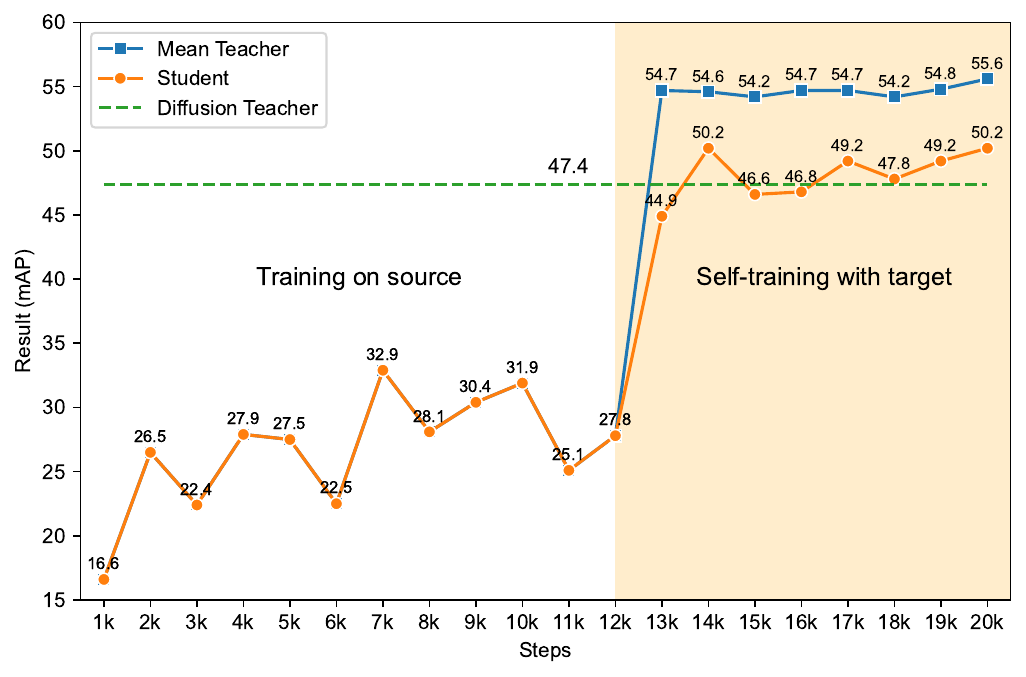}
  \caption{Evaluation results on Clipart \cite{clipart_comic_watercolor} during training. \textnormal{It is evident that the performance of the \textbf{student} significantly improves after entering self-training, even surpassing the \textbf{diffusion teacher}, and the \textbf{mean teacher} exhibits better performance compared to the student.}}
  \label{fig:training_process}
\end{figure}
%%%%%%%%%%%%%%%%%%%%%%%%%%%%%%%%%%%%%%%%%%%%%%%%%%%%%%%%%%%%%%%%%%%%%%%%%%%%%%%%%%%%%%%%%%%%%%%
%check
Object detection is a fundamental task in computer vision, with its applications permeating an array of real-world scenarios. There have been impressive strides and significant achievements in object detection, leveraging both Convolutional Neural Networks (CNNs) ~\cite{faster-rcnn,yolov3,lin2017retinanet,tian2020fcos} and transformer-based models ~\cite{carion2020Detr,zhu2021deformable}. Nonetheless, these data-driven detection algorithms wrestle with the challenging issue of domain shift: the large gap between the training data (source domain) and the testing environments (target domain) frequently results in a substantial decline in detection accuracy. This obstacle is ubiquitous across various sectors, including robotics, autonomous driving, and healthcare, and it poses a formidable barrier to the widespread applications of object detection in practice. Consequently, the deployment of domain adaptation techniques, aimed at minimizing domain disparities, has become essential to boost the robustness and generalizability of models across diverse environments.
%check

Unsupervised Domain Adaptation (UDA) methodologies have surged to the forefront of research, taking advantage of the sparse labeled data from the source domain in conjunction with copious unlabeled data from the target domain to significantly enhance cross-domain detection performance. Current UDA tactics have explored a variety of strategies including domain classifiers ~\cite{chen2018da-faster,saito2019swda,zhu2019scda,zhou2022mga,hsu2020epm}, graph matching ~\cite{li2022scan,li2022sigma,li2023sigma++}, domain randomization ~\cite{kim2019DM}, image-to-image translation ~\cite{hsu2020progressive}, and self-training frameworks ~\cite{chen2020harmonizing,roychowdhury2019automatic,li2022AT,deng2023HT}. These techniques have been crucial in achieving notable advancements in cross-domain object detection.

% check
Moreover, diffusion-based generative models ~\cite{ho2020ddpm,song2020ddim,rombach2022latent} have showcased remarkable capabilities in generating high-quality and diverse images, signaling their vast potential for a spectrum of downstream applications. Some works ~\cite{tumanyan2023plug,ddpmseg,xu2023odise} have already harnessed diffusion models for a breadth of tasks. This evidence suggests a promising avenue for employing these models to bolster cross-domain detection efficacy. Nevertheless, the step-by-step inference process of these models is not fast enough to meet the immediate processing needs of object detection. Although there has been some effort to adapt diffusion models for image generation and manipulation, as seen with tools like LoRA~\cite{lora}  and ControlNet ~\cite{controlnet}, there is a lack of research on applying diffusion models to cross-domain detection.

% check
Fortunately, current UDA methods provide us with valuable insights. Specifically, we draw inspiration from previous state-of-the-art (SOTA) approaches ~\cite{li2022AT,cao2023cmt,deng2023HT} and adopt the Mean Teacher ~\cite{mean_teacher}  self-training framework, where the teacher model generates pseudo labels for the supervised learning of student model on the target domain. The weights of the teacher model are typically updated through Exponential Moving Average (EMA) by the student model. This consistency-based self-training approach allows the student model to progressively learn from the target domain, thereby improving the performance of the detector in cross-domain detection.

% check
In our approach, we freeze all parameters of the diffusion model and extract intermediate feature from the upsampling structure of the U-Net ~\cite{unet} architecture during the inversion process. These features are then passed through a bottle-neck structure to generate hierarchical features similar to a general backbone for downstream detection tasks. This enables effective training and fine-tuning of the diffusion model with a small number of parameters, and yields discriminative feature for classification and regression tasks, leading to improved performance in cross-domain detection. The mean teacher, updated through EMA from the student model, further enhances stability and generalization.

Through the detector with the diffusion backbone for feature extraction struggles to match or surpass the performance of general backbones like ResNet ~\cite{resnet} on intra-domain. However, in the target domain, the performance of the diffusion detector surpasses them greatly. This strongly confirms the diffusion model is an incredibly powerful and highly generalized feature extractor. Furthermore, it is even more remarkable that the diffusion teacher model continues to enhance the cross-domain performance of stronger backbones, all without any increase in additional inference speed.

% check
The contributions of this paper can be summarized as follows:
\begin{itemize}
  \item We introduce a frozen-weight diffusion model as backbone, which efficiently extracts highly generalized and discriminative feature for cross-domain object detection. Notably, the diffusion-based detector, trained exclusively on the source domain, demonstrates exceptional performance when applied to the target domain.
  \item We incorporate the diffusion detector as a teacher model within the self-training framework, providing valuable guidance supervised learning of the student model on the target domain. This integration effectively enhances cross-domain detection performance without any increasing of inference time.
  \item We achieve substantial improvements in cross-domain detection. Our method achieves an average mAP improvement of 21.2\% compared with the baseline, and surpassing the current SOTA methods by 5.7\% mAP. Further experiments demonstrate that the diffusion domain teacher consistently enhances cross-domain performance for detectors with stronger backbones, leading to superior results in the target domain.
\end{itemize}

\begin{figure*}[]
   % 调整间距
  \centering
  \includegraphics[scale=0.42]{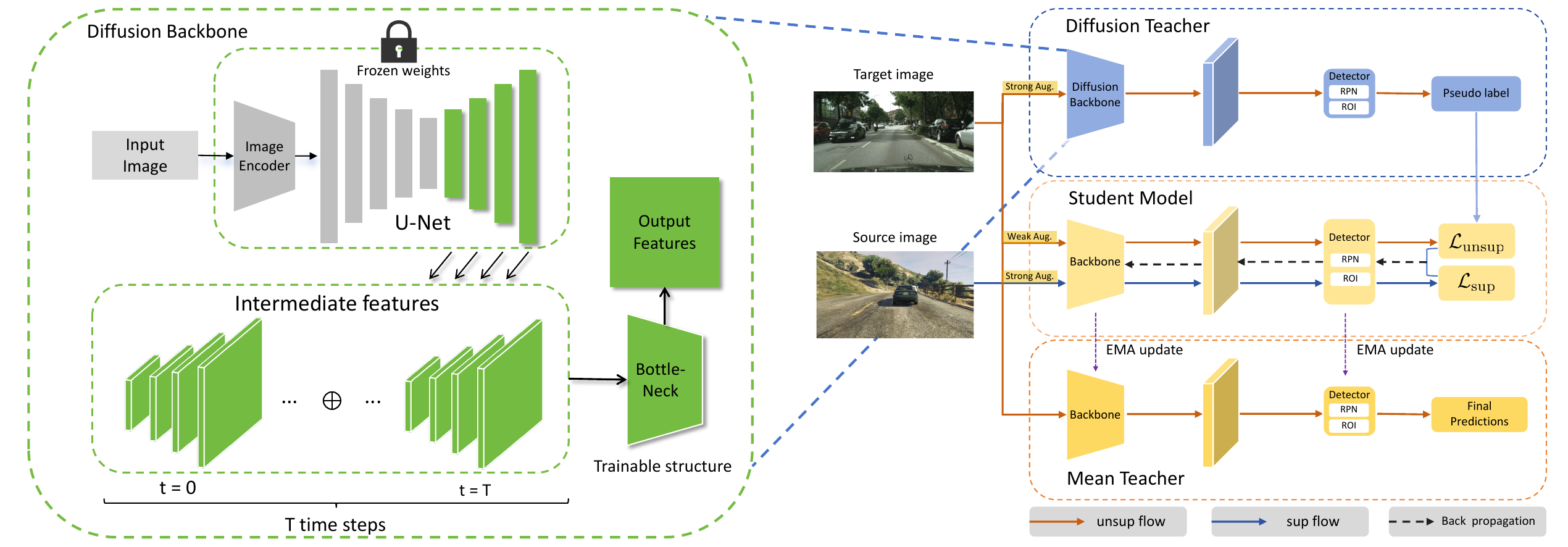}
  \caption{Overview of our proposed Diffusion Domain Teacher (DDT). \textnormal{\textbf{ Left:} We employ a frozen-weight diffusion model with bottleneck as the \textbf{diffusion backbone}, which acquires and aggregates intermediate feature from the U-Net~\cite{unet} during the inversion process at $T$ time steps for detection. \textbf{ Right:} We use the \textbf{diffusion teacher}, which is a detector with the diffusion backbone, and applied it in self-training to generate pseudo labels for unlabeled target, guiding the learning of the student. By EMA updated from the student model, \textbf{mean teacher} is refined and serves as the final model, resulting in improved cross-domain detection results.} }
  \label{fig:main_framework}
\end{figure*}

\section{Related Work}
\subsection{Object Detection}

% check
Object detection aims to locate and classify objects in given images. Deep convolutional neural networks ~\cite{resnet,VGG} have revolutionized this teak and is widely applied in real-world applications. Faster R-CNN ~\cite{faster-rcnn}, a prominent two-stage detection method, employs a region proposal network (RPN) to generate candidate regions, followed by region of interest (ROI) refinement to determine the final bounding boxes and classes. Some research is focused on improving the precision and efficiency of two-stage methods ~\cite{cascade,detectors}. In addition, researchers are investigating single-stage detectors ~\cite{lin2017retinanet,yolov3,yolov7} aimed at simplifying the detection process by integrating box regression and classification. Moreover, anchor-free detectors ~\cite{tian2020fcos,atss} that eliminate the reliance on predefined anchors have garnered significant attention. Recent research trends involve the adoption of transformer-based end-to-end detectors ~\cite{carion2020Detr,zhu2021deformable,dino}, which reconceptualize the detection task as a set prediction problem, thereby obviating the necessity for traditional handcrafted components such as anchor generation and non-maximum suppression (NMS). 

\subsection{Domain adaptation Detection}

% check
Although object detection has made significant advancements, performance can suffer greatly due to domain shifts between training and test data. To address this issue, UDA aims to mitigate the impact of domain shifts by leveraging labeled source data and unlabeled target data. Initially, studies inspired by GANs ~\cite{gan} introduced domain adversarial training ~\cite{DANN}, which minimized domain gaps by extracting invariant feature. This approach is later adapted to detection tasks ~\cite{chen2018da-faster,saito2019swda,zhu2019scda,zhou2022mga,hsu2020epm}, resulting in notable improvements in performance on the target domain. Some methods ~\cite{hsu2020progressive} focus on reducing inter-domain differences at image level, by applying image-to-image translate like CycleGAN ~\cite{cyclegan}.

%check
Recently, self-training domain adaptation methods ~\cite{umt,li2022AT,deng2023HT,cao2023cmt} have achieved better results in cross-domain object detection by optimizing the learning of pseudo labels in the target domain. For example, AT ~\cite{li2022AT} combines domain adversarial training and self-training to improve the quality of pseudo labels. UMT ~\cite{umt} utilizes consistency learning for a teacher model to generate high-quality pseudo labels to improve cross-domain detection. CMT ~\cite{cao2023cmt} introduces contrastive learning to optimize the utilization and alignment of features from the source and target domain. HT ~\cite{deng2023HT} optimizes the generation of pseudo labels by applying consistency measures in regression and classification. Overall, self-training methods in cross-domain object detection enhance the detection performance in the target domain by improving the quality of pseudo labels, with the teacher model being updated from the student model. 

\subsection{Diffusion Models}

Diffusion models ~\cite{ho2020ddpm,song2020ddim,rombach2022latent,dalle,imagen} have achieved impressive results in image generation, surpassing previous models like GAN ~\cite{gan}. With their strong generative and generalization capabilities, some research has begun to explore the potential of diffusion models in feature representation and their application to downstream tasks. For instance, DDPMSeg ~\cite{ddpmseg} and ODISE ~\cite{xu2023odise} utilize feature extracted from diffusion models for semantic and panoptic segmentation tasks, respectively. DIFT ~\cite{tang2023emergent} and HyperFeature ~\cite{hyperfeatures} use the diffusion model to discover correspondences in images. This inspires us to consider the application of diffusion models for improving cross-domain detection tasks.

\section{Approach}

In this section, we present our Diffusion Domain Teacher (DDT) framework in detail. First, in Sec. ~\ref{sec:3.1}, we review the formulation of Unsupervised Domain Adaptation Detection (UDAD). Then, in Sec. ~\ref{sec:3.2}, we provide a detailed description of how the frozen-weight diffusion model serves as a feature extractor, producing hierarchical features, to adapt to the detection task. Furthermore, in Sec. ~\ref{sec:3.3}, we explain the application of the diffusion teacher detector in thr self-training framework, where pseudo labels generated on the unlabeled target domain guide the supervised learning of the student model. Finally, we summarize the total training objective.

\subsection{Formulation of Unsupervised Domain adaptation Detection} \label{sec:3.1}

% check
To be specific, we denote a given set of $N_s$ samples $\mathcal{S}=\left\{X_s^i, Y_s^i\right\}_{i=1}^{N_s}$ as source domian, where $X_s^i$ represents an image and $Y_s^i$ represents the bounding box with category labels in the respective image. Similarly, we denote the target domain data as $\mathcal{T}=\left\{X_t^i\right\}_{i=1}^{N_s}$, which consists of $N_t$ unlabeled samples. Exactly, the distribution of $\mathcal{S}$ and $\mathcal{T}$, including the distributions of images from $P(X_s)$ and $P(X_t)$ (e.g., style, scene, weather), labels $P(Y_s)$ and $P(Y_t)$ (e.g., the shapes, sizes, and density of instance), and even the scales of $N_s$ and $N_t$  are different, denoted as $P(\mathcal{S}) \ne P(\mathcal{T})$, is what we refer to as a cross-domain detection problem. Furthermore, relying solely on supervised learning from the labeled source domain results in an inherent bias towards source domain in cross-domain detection. Domain adaptation for detection aims to improving the performance on the target domain by reducing the dissimilarity between $\mathcal{S}$ and $\mathcal{T}$, seeking to a domain-invariant detector.

\subsection{Fozen-Diffusion Feature Extractor} \label{sec:3.2}

% check
Diffusion generative models ~\cite{ho2020ddpm,song2020ddim,rombach2022latent} aim to minimize the discrepancy between the distribution of images generated by the model, denoted as $P_{\theta}(x)$, and the distribution of the training data, denoted as $P_{\text{data}}(x)$. During training, gaussian noise of varying magnitudes is added to the clean training data, commonly referred to as \textit{diffusion} process. The diffusion process starts with a clean image $x_0$ from the training data and generates a noisy image $x_t$ by mixing $x_0$ with noise of different magnitudes:
\begin{equation}
  x_t=\sqrt{\bar{\alpha}_t} x_0+\sqrt{1-\bar{\alpha}_t} \epsilon
\end{equation}

%check
where $\epsilon \sim \mathcal{N}(0, \mathbf{I})$ represents randomly sampled noise, and $t \in [0, T]$ denotes the time step, where larger values correspond to adding more noise. The amount of noise added is determined by ${\alpha_t}$, which is a predefined noise schedule, and $\bar{\alpha}_t = \alpha_1 \alpha_2 \ldots \alpha_t$. The model $f_{\theta}$ is trained to predict the input noise $\epsilon$, given $x_t$ and $t$, typically using structures like U-Net ~\cite{unet}.

% check
The iterative process of the diffusion model poses challenges when directly applied to downstream supervised tasks. We extract intermediate feature at a specific time step $t$ during the inversion process, and apply these features for regression and classification tasks in the detection task. Specifically, we append a input noise corresponding the time step $t$ to the input image, shift it to $x_t$ and then input it along with $t$ into $f_{\theta}$ to extract activation layers as intermediate feature. More specifically, we apply the intermediate feature from the four stages of the upsampling process in the denoise network U-Net ~\cite{unet}. For each input image, we concatenate multiple time step feature together and employ a bottle-neck structure to project the feature into hierarchical layers with a channel size of $[256, 512, 1024, 2048]$, similar to the output of ResNet ~\cite{resnet}, which is directly applied to the object detection task, as shown in the left side of Fig. ~\ref{fig:main_framework}.

\subsection{Diffusion Teacher Guided Self-training Framework} \label{sec:3.3}

% check
We employ a detector that extracts feature using the diffusion model and trained on the source domain as the teacher model, denoted as $\mathcal{F}_{\text{diff}}$. It is used to generate pseudo labels $\bar{Y}_{t}$ on the target domain $\mathcal{T}$, where $\bar{Y}_{t} = \mathcal{F}_{diff}(X_t)$. These pseudo labels are constructed to form a new dataset $\mathcal{\bar{T}}=\left\{X_t^i, \bar{Y}_t^i\right\}_{i=1}^{N_t}$. Subsequently, we optimize the student model using the pseudo labels. We introduce a hyperparameter $\sigma$ as a threshold for the confidence scores of the output for the teacher model, enabling us to select more reliable pseudo labels.

% check
We define the supervised learning of the student model $\mathcal{F}_{\text{stu}}$ on the source domain as follows:
\begin{equation}
  \begin{aligned}
    \mathcal{L}_{\text{sup}}\left(X_s, Y_s\right)= & \mathcal{L}_{\text{cls}}^{\text{RPN}}\left(X_s, Y_s\right)+\mathcal{L}_{\text{reg}}^{\text{RPN}}\left(X_s, Y_s\right) \\
                                                   & +\mathcal{L}_{\text{cls}}^{\text{ROI}}\left(X_s,Y_s\right)+\mathcal{L}_{\text{reg}}^{\text{ROI}}\left(X_s, Y_s\right)
  \end{aligned}
\end{equation}

%check
where RPN is used to generate potential candidate regions, and ROI performs classification and bounding box regression on these candidate regions to obtain more accurate class and bounding box predictions, denoted as $\text{cls}$ and $\text{reg}$, respectively.  Similarly, we define the learning of the student model in the target domain as follows:
\begin{equation}
  \begin{aligned}
    \mathcal{L}_{\text{unsup}}\left(X_t, \bar{Y}_t\right)= & \mathcal{L}_{\text{cls}}^{\text{RPN}}\left(X_t, \bar{Y}_t\right)+\mathcal{L}_{\text{reg}}^{\text{RPN}}\left(X_t, \bar{Y}_t\right) \\
                                                           & +\mathcal{L}_{\text{cls}}^{\text{ROI}}\left(X_t,\bar{Y}_t\right)+\mathcal{L}_{\text{reg}}^{\text{ROI}}\left(X_t, \bar{Y}_t\right)
  \end{aligned}
\end{equation}

%check
Then, we employ EMA to update a mean teacher model $\mathcal{F}_{mean}$ by copying the weights from the student model. We define this process as follows:
\begin{equation}
  \theta_t \leftarrow \alpha \theta_t+(1-\alpha) \theta_s
\end{equation}

%check
where $t$ and $s$ represent the parameters of $\mathcal{F}_{\text{mean}}$ and $\mathcal{F}_{\text{stu}}$, respectively. By employing EMA to update the mean teacher model, we aim to create a more stable and robust model by gradually incorporating the knowledge learned by the student model over time. We select the output of $\mathcal{F}_{\text{mean}}$ as result for predicting.

%check
We apply a hyper parameter $\lambda$ to adjust the weights between $\mathcal{L}_{\text{unsup}}$ and $\mathcal{L}_{\text{sup}}$ . The final formulation of our comprehensive loss function is summarized as follows:
\begin{equation}
  \mathcal{L}=\mathcal{L}_{\text{sup}}+\lambda\cdot \mathcal{L}_{\text {unsup}}
\end{equation}

%check
In our DDT framework, following ~\cite{li2022AT}, we employ \textit{Weak Augmentation} to provide target domain images to the diffusion teacher model for generating reliable and accurate pseudo labels. Simultaneously, we apply \textit{Strong Augmentation} to the images as inputs to the student model, as illustrated in Fig. ~\ref{fig:main_framework}. Specifically, \textit{Weak Augmentation} includes random crop and random horizontal flip, while \textit{Strong Augmentation} involves color transformations such as color space conversion, contrast adjustment, equalization, sharpness enhancement, and posterization, as well as spatial transformations such as rotation, shear, and translation of the position.

%check
%%%%%%%%%%%%%%%%%%%%%%%%%%%%%%%%%%%%%%%%%%%%%%%%%%%%%%%%%%%%%%%%%%%%%%%%%%%%%%%%%%%%%%%%%%%%%%%%%%%%%%%%%%%%%%%
\begin{figure*}[h]
  \centering
  \includegraphics[width=0.8\linewidth]{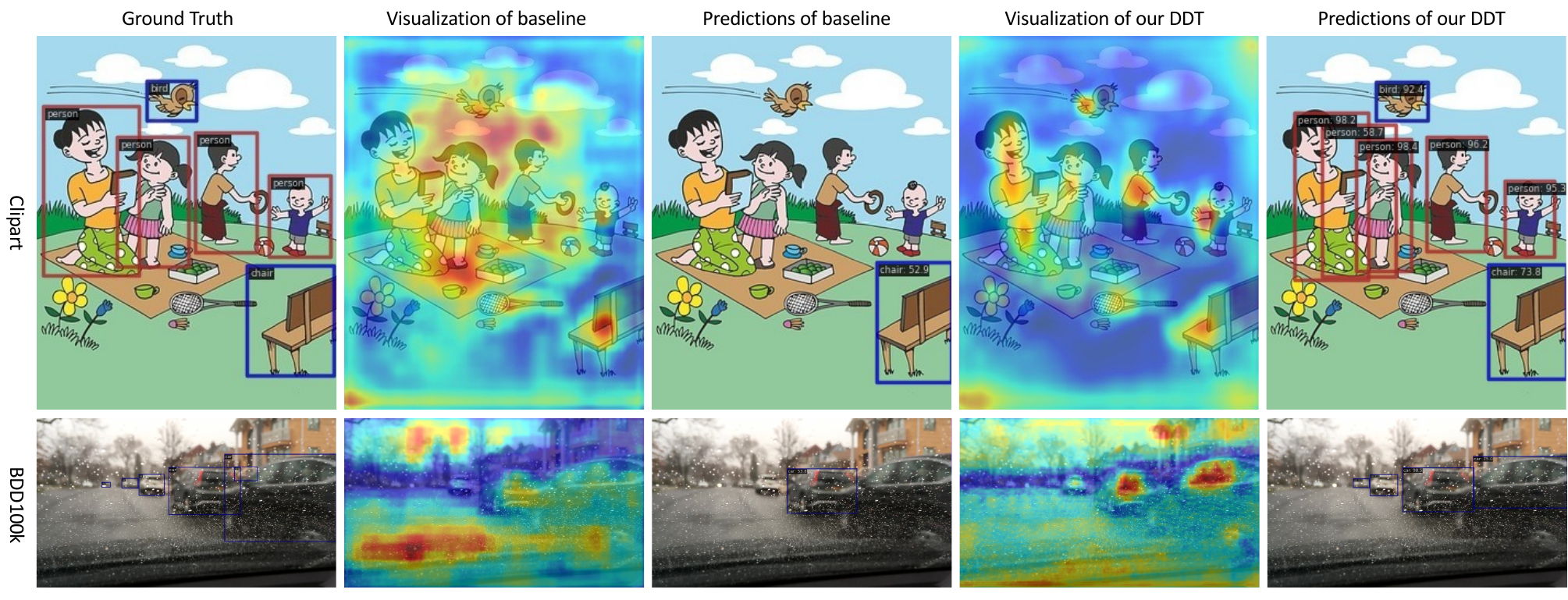}
  \caption{Qualitative prediction results and feature visualization of baseline and our DDT. \textnormal{Compared to the baseline, our method focuses more on specific classes of objects in the target domain images, effectively reducing the numbers of false negative on Clipart ~\cite{cityscapes} (\textbf{first row}) and BDD100K ~\cite{bdd100k} (\textbf{second row}).}}
  \label{fig:prediction result}
\end{figure*}
%%%%%%%%%%%%%%%%%%%%%%%%%%%%%%%%%%%%%%%%%%%%%%%%%%%%%%%%%%%%%%%%%%%%%%%%%%%%%%%%%%%%%%%%%%%%%%%%%%%%%%%%%%%%%%%

\section{Experiments}
\subsection{Datasets}

\quad \textbf{Cityscapes.}\  Cityscapes ~\cite{cityscapes} dataset provides a diverse of urban scenes from 50 cities. It includes 2,975 training images and 500 validation images with detailed annotations. The dataset covers 8 detection categories, using bounding boxes sourced from instance segmentation.

\textbf{BDD100K.}\ BDD100K ~\cite{bdd100k} dataset is a comprehensive collection of 100,000 images specifically designed for autonomous driving applications. The dataset offers detailed detection annotations with 10 categories.

\textbf{Sim10K.}\ Sim10k ~\cite{sim10k} is a synthetic dataset comprising 10,000 rendered images simulated within the Grand Theft Auto gaming engine, specifically designed to facilitate the training and evaluation of object detection algorithms in autonomous driving systems.

\textbf{VOC.}\ VOC ~\cite{voc} is a general-purpose object detection dataset that includes bounding box and class annotations for common objects across 20 categories from the real world. Following ~\cite{li2022AT}, we combined the PASCAL VOC 2007 and 2012 editions, resulting in a total of 16,551 images.

\textbf{Clipart.}\ Clipart ~\cite{clipart_comic_watercolor} dataset comprises 1,000 clipart images across the same 20 categories as the VOC dataset, exhibiting significant differences from real-world images. Following ~\cite{li2022AT}, we utilize 500 images each for training and testing purposes.

\textbf{Comic.}\ Comic ~\cite{clipart_comic_watercolor} dataset consists of 2,000 comic-style images, featuring 6 categories shared with the VOC dataset. Following ~\cite{dadapt}, we allocate 1,000 images each for training and testing.

\textbf{Watercolor.}\ Watercolor ~\cite{clipart_comic_watercolor} dataset contains 2,000 images in a watercolor painting style, with 6 categories shared with the VOC dataset. Following ~\cite{li2022AT}, we use 1,000 images for both training and testing.

\subsection{Cross-domian Detection Settings}

\quad \textbf{Cross-Camera.}\ We train on Cityscapes ~\cite{cityscapes} (source domain) and validate on BDD100K ~\cite{bdd100k} (target domain) to evaluate the cross-camera detection performance in diverse weather and scene conditions. We focus on the 7 same categories as SWDA ~\cite{saito2019swda}.

\textbf{Synthetic to Real (Syn2Real).} \ We train on Sim10K (source domain) and validate on Cityscapes ~\cite{cityscapes} and BDD100K ~\cite{bdd100k} (target domian)to validate the performance of synthetic-to-real detection. Following SWDA ~\cite{saito2019swda}, we focus on the shared category \textit{car}.

\textbf{Real to Artistic.}\ We train on the VOC ~\cite{voc} (source domain) and perform validation on the Clipart ~\cite{clipart_comic_watercolor}, Comic ~\cite{clipart_comic_watercolor}, and Watercolor ~\cite{clipart_comic_watercolor} (target domains) to assess cross-domain detection performance from real-world images to artistic styles. Referring to AT ~\cite{li2022AT} and D-ADAPT ~\cite{dadapt}, we respectively apply the 20, 6, and 6 shared categories between VOC and each of the Clipart, Comic, and Watercolor.

\subsection{Implementation Details}
Following ~\cite{saito2019swda,li2022AT,dadapt}, we use Faster R-CNN ~\cite{faster-rcnn} as the default detector with a ResNet101 ~\cite{resnet} backbone pretrained on ImageNet ~\cite{imagenet}, implemented with MMDetection ~\cite{mmdetection}. The training and testing sizes of images are set to (1333, 800) for Cityscapes, BDD100K, and Sim10K, and (1200, 600) for VOC, Clipart, Comic, and Watercolor. The models are trained with 20,000 steps on two 3090 GPUs, with a total batch size of 16. We employ the SGD optimizer with an initial learning rate of 0.02, following the default settings in MMDetection.

In self-training, we refer to the settings in ~\cite{li2022AT,deng2023HT} to apply both weak and strong augmentation on the unlabeled target domain. We employ the EMA update parameter $\alpha$ of 0.999 for the mean teacher model updates and simply set the loss weight $\lambda$ to 1. We train exclusively on the source domain for the first 12000 steps and then perform joint training on both the source and target domain for the remaining 8000 steps.

For evaluation, we report the Average Precision (AP) for each object category and the mean Average Precision (mAP) across all categories, with applying an Intersection over Union (IoU) threshold of 0.5.

\subsection{Results and Comparisons}

In this section, we present the evaluation result of our DDT framework along with other SOTA approaches. Current cross-domain object detection methods employ different detectors ~\cite{faster-rcnn,tian2020fcos,SSD,zhu2021deformable}, which we refer to as FRCNN, FCOS, SSD, and DDETR in our table. Furthermore, the backbones with varying depths, including ResNet-18, ResNet-50, ResNet-101 ~\cite{resnet}, and VGG-16 ~\cite{VGG}, are denoted as R18, R50, R101, and V16, respectively. To provide a comprehensive comparison, we report the results of our method with ResNet18, ResNet50, and ResNet101. The \textit{baseline} refers to the results that only train on the source domain and test on the target domain.

%%%%%%%%%%%%%%%%%%%%%%%%%%%%%%%%%%%%%%%%%%%%

\begin{table}[]
  \renewcommand{\arraystretch}{1.15}
  \setlength{\tabcolsep}{1pt}
  \centering
  \caption{Quantitative results on adaptation from \small \textbf{Cityscapes to BDD100K (Cs$\rightarrow$B)}. The bold indicates the best results.}
  \label{tab:city_to_bdd100k}
  \resizebox{\columnwidth}{!}{%
    \begin{tabular}{ccc|ccccccc|c}
      \toprule
      Method                                       & Reference                      & Detector                          & bicycle & bus  & car  & mcycle & person & rider & truck & mAP                                                \\ \midrule
      \textbf{DA-Faster} ~\cite{chen2018da-faster} & \small{\textit{CVPR'18}}  & FRCNN\footnotesize -V16                   & 22.4    & 18.0 & 44.2 & 14.2   & 28.9   & 27.4  & 19.1  & 24.9                                               \\
      \textbf{SWDA} ~\cite{saito2019swda}          & \small{\textit{CVPR'19}}  & FRCNN\footnotesize -V16                   & 23.1    & 20.7 & 44.8 & 15.2   & 29.5   & 29.9  & 20.2  & 26.2                                               \\
      \textbf{SCDA} ~\cite{zhu2019scda}            & \small{\textit{CVPR'19}}  & FRCNN\footnotesize -V16                   & 23.2    & 19.6 & 44.4 & 14.8   & 29.3   & 29.2  & 20.3  & 25.8                                               \\
      \textbf{CRDA} ~\cite{crda}                   & \small{\textit{CVPR'20}}  & FRCNN\footnotesize -R101                  & 25.5    & 20.6 & 45.8 & 14.9   & 32.8   & 29.3  & 22.7  & 27.4                                               \\
      \textbf{SED} ~\cite{sed}                     & \small{\textit{AAAI'21}}  & FRCNN\footnotesize -V16                   & 25.0    & 23.4 & 50.4 & 18.9   & 32.4   & 32.6  & 20.6  & 29.0                                               \\
      \textbf{TDD} ~\cite{tdd}                     & \small{\textit{CVPR'22}}  & FRCNN\footnotesize -V16                   & 28.8    & 25.5 & 53.9 & 24.5   & 39.6   & 38.9  & 24.1  & 33.6                                               \\
      \textbf{PT} ~\cite{pt}                       & \small{\textit{ICML'22}}  & FRCNN\footnotesize -V16                   & 28.8    & 33.8 & 52.7 & 23.0   & 40.5   & 39.9  & 25.8  & 34.9                                               \\
      \textbf{EPM} ~\cite{hsu2020epm}              & \small{\textit{ECCV'20}}  & FCOS\footnotesize -R101                   & 20.1    & 19.1 & 55.8 & 14.5   & 39.6   & 26.8  & 18.8  & 27.8                                               \\
      \textbf{SIGMA} ~\cite{li2022sigma}           & \small{\textit{CVPR'22}}  & FCOS\footnotesize -R50                    & 26.3    & 23.6 & 64.1 & 17.9   & 46.9   & 29.6  & 20.2  & 32.7                                               \\
      \textbf{SIGMA++} ~\cite{li2023sigma++}       & \small{\textit{TPAMI'23}} & FRCNN\footnotesize -V16                   & 27.1    & 26.3 & 65.6 & 17.8   & 47.5   & 30.4  & 21.1  & 33.7                                               \\
      \textbf{NSA} ~\cite{nsa}                     & \small{\textit{ICCV'23}}  & FRCNN\footnotesize -V16                   & /       & /    & /    & /      & /      & /     & /     & 35.5                                               \\
      \textbf{HT} ~\cite{deng2023HT}               & \small{\textit{CVPR'23}}  & FCOS\footnotesize -V16                    & 38.0    & 30.6 & 63.5 & 28.2   & 53.4   & 40.4  & 27.4  & 40.2                                               \\ \midrule \midrule
      Baseline                                     & \multirow{2}{*}{/}             & \multirow{2}{*}{FRCNN\footnotesize -R18}  & 23.8    & 13.0 & 51.8 & 17.0   & 42.5   & 27.4  & 15.7  & 27.3                                               \\
      \textbf{DDT(Ours)}                           &                                &                                   & 36.8    & 27.0 & 64.9 & 25.8   & 55.3   & 39.2  & 27.3  & 39.5\footnotesize \textcolor{red}{\textbf{+12.2}}          \\
      Baseline                                     & \multirow{2}{*}{/}             & \multirow{2}{*}{FRCNN\footnotesize -R50}  & 24.8    & 16.5 & 53.9 & 15.4   & 45.3   & 27.6  & 18.2  & 28.8                                               \\
      \textbf{DDT(Ours)}                           &                                &                                   & 39.0    & 31.6 & 65.9 & 30.2   & 57.7   & 39.8  & 28.6  & 41.8\footnotesize \textcolor{red}{\textbf{+13.0}}          \\
      Baseline                                     & \multirow{2}{*}{/}             & \multirow{2}{*}{FRCNN\footnotesize -R101} & 25.9    & 18.4 & 48.8 & 17.2   & 41.1   & 29.8  & 21.7  & 29.0                                               \\
      \textbf{DDT(Ours)}                           &                                &                                   & 40.3    & 32.3 & 66.7 & 31.8   & 59.1   & 41.6  & 31.8  & \textbf{43.4}\footnotesize \textcolor{red}{\textbf{+14.4}} \\ \bottomrule
    \end{tabular}%
  }
\end{table}
%%%%%%%%%%%%%%%%%%%%%%%%%%%%%%%%%%%%%%%%%%%%%%%

%Cross-camera settings的结果在表中。与Source only的结果相比,我们分别在ResNet18、Resnet50,resnet101上分别获得了12.2,13.0和14.4的提升。我们最好的结果在target domian上获得了43.4mAp,高出之前的SOTA方法HT3.2mAP,并大幅度领先于其他方法。我们可以发现,采用自训练架构的PT、HT方法,通过改善生成的伪标签的质量,已经取得了巨大的性能提升。我们的DDT方法借助于Diffusion模型的强大特征表示能力和在多样性图像上的优异表现,通过简单的预测蒸馏和一致性训练,优化目标域伪标签的生成,最终显著改善了cross-camera的检测表现。

\textbf{Cross-camera adaptation.} Tab. ~\ref{tab:city_to_bdd100k} presents the results of the Cross-camera settings. Our DDT method achieved the best performance with mAP 43.4 on the target domain, surpassing the previous SOTA method HT ~\cite{deng2023HT} by 3.2 mAP and outperforming other methods by a significant margin. Notably, AT ~\cite{li2022AT} and HT ~\cite{deng2023HT} utilize self-training framework, have demonstrated substantial performance improvements by enhancing the quality of generated pseudo labels. Leveraging the powerful feature representation capability of the diffusion model and its exceptional performance on diverse images, our DDT method achieve a remarkable enhancement in cross-camera detection.

\begin{table}[]
   % 调整间距
  \setlength{\tabcolsep}{4pt}
  \centering
  \caption{Quantitative results on adaptation from \small \textbf{Sim10K to BDD100K (S$\rightarrow$B)}. \normalsize The bold indicates the best results.}
  \label{tab:sim10k_to_bdd100k}
  \begin{tabular}{ccc|c}
    \toprule
    Method                              & Reference                     & Detector                          & mAP(car)                                           \\ \midrule
    \textbf{SWDA} ~\cite{saito2019swda} & \small{\textit{CVPR'19}} & FRCNN\footnotesize -V16                   & 42.9                                               \\
    \textbf{CDN} ~\cite{cdn}            & \small{\textit{ECCV'20}} & FRCNN\footnotesize -V16                   & 45.3                                               \\ \midrule \midrule
    Baseline                            & \multirow{2}{*}{/}            & \multirow{2}{*}{FRCNN\footnotesize -R18}  & 30.9                                               \\
    \textbf{DDT(Ours)}                  &                               &                                   & 57.2\footnotesize \textcolor{red}{\textbf{+26.3}}          \\
    Baseline                            & \multirow{2}{*}{/}            & \multirow{2}{*}{FRCNN\footnotesize -R50}  & 34.4                                               \\
    \textbf{DDT(Ours)}                  &                               &                                   & 57.6\footnotesize \textcolor{red}{\textbf{+23.2}}          \\
    Baseline                            & \multirow{2}{*}{/}            & \multirow{2}{*}{FRCNN\footnotesize -R101} & 34.2                                               \\
    \textbf{DDT(Ours)}                  &                               &                                   & \textbf{58.3}\footnotesize \textcolor{red}{\textbf{+24.1}} \\ \bottomrule
  \end{tabular}
\end{table}

\begin{table}[]
   % 调整间距
  \setlength{\tabcolsep}{4pt}
  \centering
  \caption{Quantitative results on adaptation from Sim10K to Cityscapes (S$\rightarrow$Cs). The bold indicates the best results.}
  \label{tab:sim10k_to_city}
  \begin{tabular}{ccc|c}
    \toprule
    Method                                  & Reference                        & Detector                          & mAP(car)                                  \\ \midrule
    % \textbf{DA-Faster} ~\cite{chen2018da-faster} & \small{\textit{CVPR'18}}    & FRCNN\footnotesize -V16                   & 39.0                                      \\
    % \textbf{SWDA} ~\cite{saito2019swda}          & \small{\textit{CVPR'19}}    & FRCNN\footnotesize -V16                   & 40.7                                      \\
    % \textbf{HTCN} ~\cite{htcn}                   & \small{\textit{CVPR'20}}    & FRCNN\footnotesize -R101                  & 42.5                                      \\
    % \textbf{UMT} ~\cite{umt}                     & \small{\textit{CVPR'21}}    & FRCNN\footnotesize -R101                  & 43.1                                      \\
    \textbf{SSAL} ~\cite{ssal}              & \small{\textit{NeurIPS'22}} & FCOS\footnotesize -R50                    & 51.8                                      \\
    \textbf{O2NET} ~\cite{o2net}            & \small{\textit{ACMMM'22}}   & DDETR\footnotesize -R50                   & 54.1                                      \\
    \textbf{DDF} ~\cite{ddf}                & \small{\textit{TMM'22}}     & FRCNN\footnotesize -R50                   & 44.3                                      \\
    \textbf{D-ADAPT} ~\cite{dadapt}         & \small{\textit{ICLR'22}}    & FRCNN\footnotesize -R50                   & 51.9                                      \\
    \textbf{SCAN} ~\cite{scan}              & \small{\textit{AAAI'22}}    & FCOS\footnotesize -V16                    & 52.6                                      \\
    \textbf{MTTrans} ~\cite{MTTrans}        & \small{\textit{ECCV'22}}    & DDETR\footnotesize -R50                   & 57.9                                      \\
    \textbf{SIGMA} ~\cite{li2022sigma}      & \small{\textit{CVPR'22}}    & FCOS\footnotesize -R50                    & 53.7                                      \\
    \textbf{TDD}  ~\cite{li2023sigma++}     & \small{\textit{CVPR'22}}    & FRCNN\footnotesize -V16                   & 53.4                                      \\
    \textbf{MGA}  ~\cite{zhou2022mga}       & \small{\textit{CVPR'22}}    & FCOS\footnotesize -R101                   & 54.1                                      \\
    \textbf{OADA}  ~\cite{oada}             & \small{\textit{ECCV'22}}    & FCOS\footnotesize -V16                    & 59.2                                      \\
    \textbf{SIGMA++}  ~\cite{li2023sigma++} & \small{\textit{TPAMI'23}}   & FCOS\footnotesize -V16                    & 53.7                                      \\
    \textbf{CIGAR} ~\cite{CIGAR}            & \small{\textit{CVPR'23}}    & FCOS\footnotesize -V16                    & 58.5                                      \\
    \textbf{NSA} ~\cite{nsa}                & \small{\textit{ICCV'23}}    & FRCNN\footnotesize -V16                   & 56.3                                      \\
    \textbf{HT} ~\cite{deng2023HT}          & \small{\textit{CVPR'23}}    & FRCNN\footnotesize -V16                   & \textbf{65.5}                             \\ \midrule \midrule
    Baseline                                & \multirow{2}{*}{/}               & \multirow{2}{*}{FRCNN\footnotesize -R18}  & 42.9                                      \\
    \textbf{DDT(Ours)}                      &                                  &                                   & 62.3\footnotesize \textcolor{red}{\textbf{+19.4}} \\
    Baseline                                & \multirow{2}{*}{/}               & \multirow{2}{*}{FRCNN\footnotesize -R50}  & 43.0                                      \\
    \textbf{DDT(Ours)}                      &                                  &                                   & 62.7\footnotesize \textcolor{red}{\textbf{+19.7}} \\
    Baseline                                & \multirow{2}{*}{/}               & \multirow{2}{*}{FRCNN\footnotesize -R101} & 43.4                                      \\
    \textbf{DDT(Ours)}                      &                                  &                                   & 64.0\footnotesize \textcolor{red}{\textbf{+20.6}} \\ \bottomrule
  \end{tabular}%
\end{table}

\textbf{{Synthetic to Real adaptation.}} In Tab. ~\ref{tab:sim10k_to_bdd100k}, our method achieves improvements of 26.3, 23.3, and 24.2 mAP on BDD100K ~\cite{bdd100k} compared with baseline, respectively, surpassing the results of previous algorithms SWDA ~\cite{saito2019swda} and CDN ~\cite{cdn}. Similarly, our method obtains improvements of 19.4, 19.7, and 19.3 mAP on Cityscapes ~\cite{cityscapes}, respectively, surpassing all methods except HT ~\cite{deng2023HT} in Table ~\ref{tab:sim10k_to_city}.
It can be observed that detectors for synthetic-to-real detection, due to the significant differences between synthetic and real-world images, does not perform well with source data only, while our method significantly improves the cross-domain performance from Sim10K ~\cite{sim10k} to BDD100K ~\cite{bdd100k} and Cityscapes ~\cite{cityscapes}.

% \usepackage{multirow}
% \usepackage{graphicx}
\begin{table*}[]
  \renewcommand{\arraystretch}{1.15}
   % 调整间距
  \setlength{\tabcolsep}{3pt}
  \centering
  \caption{Quantitative results on adaptation from VOC to Clipart (V$\rightarrow$Ca). The bold indicates the best results.}
  \label{tab:voc_to_clipart}
  \resizebox{\textwidth}{!}{%
    \begin{tabular}{ccc|cccccccccccccccccccc|c}
      \toprule
      Method                           & Reference                     & Detector                          & aero & bcycle & bird & boat & bottle & bus  & car  & cat  & chair & cow  & table & dog  & horse & bike & psn  & plant & sheep & sofa & train & tv   & mAP                                       \\ \midrule
      \textbf{AT} ~\cite{li2022AT}     & \small{\textit{CVPR'22}} & FRCNN\footnotesize -V16                   & 33.8 & 60.9   & 38.6 & 49.4 & 52.4   & 53.9 & 56.7 & 7.5  & 52.8  & 63.5 & 34.0  & 25.0 & 62.2  & 72.1 & 77.2 & 57.7  & 27.2  & 52.0 & 55.7  & 54.1 & 49.3                                      \\
      \textbf{D-ADAPT} ~\cite{dadapt}  & \small{\textit{ICLR'22}} & FRCNN\footnotesize -R50                   & 56.4 & 63.2   & 42.3 & 40.9 & 45.3   & 77.0 & 48.7 & 25.4 & 44.3  & 58.4 & 31.4  & 24.5 & 47.1  & 75.3 & 69.3 & 43.5  & 27.9  & 34.1 & 60.7  & 64.0 & 49.0                                      \\
      \textbf{TIA} ~\cite{tia}         & \small{\textit{CVPR'22}} & FRCNN\footnotesize -R101                  & 42.2 & 66.0   & 36.9 & 37.3 & 43.7   & 71.8 & 49.7 & 18.2 & 44.9  & 58.9 & 18.2  & 29.1 & 40.7  & 87.8 & 67.4 & 49.7  & 27.4  & 27.8 & 57.1  & 50.6 & 46.3                                      \\
      \textbf{LODS} ~\cite{lods}       & \small{\textit{CVPR'22}} & FRCNN\footnotesize -R101                  & 43.1 & 61.4   & 40.1 & 36.8 & 48.2   & 45.8 & 48.3 & 20.4 & 44.8  & 53.3 & 32.5  & 26.1 & 40.6  & 86.3 & 68.5 & 48.9  & 25.4  & 33.2 & 44.0  & 56.5 & 45.2                                      \\
      \textbf{CIGAR} ~\cite{CIGAR}     & \small{\textit{CVPR'23}} & FCOS\footnotesize -R101                   & 35.2 & 55.0   & 39.2 & 30.7 & 60.1   & 58.1 & 46.9 & 31.8 & 47.0  & 61.0 & 21.8  & 26.7 & 44.6  & 52.4 & 68.5 & 54.4  & 31.3  & 38.8 & 56.5  & 63.5 & 46.2                                      \\
      \textbf{CMT}  ~\cite{cao2023cmt} & \small{\textit{CVPR'23}} & FRCNN\footnotesize -V16                   & 39.8 & 56.3   & 38.7 & 39.7 & 60.4   & 35.0 & 56.0 & 7.1  & 60.1  & 60.4 & 35.8  & 28.1 & 67.8  & 84.5 & 80.1 & 55.5  & 20.3  & 32.8 & 42.3  & 38.2 & 47.0                                      \\ \midrule \midrule
      Baseline                         & \multirow{2}{*}{/}            & \multirow{2}{*}{FRCNN\footnotesize -R18}  & 25.0 & 36.4   & 16.2 & 19.6 & 34.3   & 50.7 & 30.3 & 0.2  & 33.6  & 5.5  & 22.1  & 6.5  & 23.3  & 47.9 & 36.1 & 26.8  & 3.2   & 18.5 & 31.7  & 25.1 & 24.6                                      \\
      \textbf{DDT(Ours)}               &                               &                                   & 55.8 & 64.8   & 37.0 & 37.3 & 46.5   & 61.7 & 52.4 & 3.3  & 50.8  & 39.9 & 36.6  & 25.1 & 46.9  & 87.4 & 77.6 & 52.7  & 25.2  & 40.1 & 45.0  & 44.9 & 46.5\footnotesize \textcolor{red}{\textbf{+21.9}} \\
      Baseline                         & \multirow{2}{*}{/}            & \multirow{2}{*}{FRCNN\footnotesize -R50}  & 29.5 & 36.1   & 22.8 & 25.6 & 34.2   & 41.3 & 27.6 & 3.4  & 39.2  & 8.3  & 28.0  & 5.2  & 18.8  & 47.9 & 34.1 & 37.6  & 4.0   & 21.8 & 31.8  & 24.0 & 26.0                                      \\
      \textbf{DDT(Ours)}               &                               &                                   & 48.9 & 64.9   & 41.1 & 48.1 & 60.5   & 76.6 & 60.2 & 7.9  & 57.4  & 51.5 & 40.8  & 33.0 & 55.4  & 98.0 & 82.8 & 60.1  & 35.2  & 36.0 & 50.8  & 55.8 & 53.2\footnotesize \textcolor{red}{\textbf{+27.2}} \\
      Baseline                         & \multirow{2}{*}{/}            & \multirow{2}{*}{FRCNN\footnotesize -R101} & 36.7 & 27.2   & 22.6 & 22.8 & 38.5   & 46.4 & 32.2 & 10.7 & 40.7  & 7.2  & 27.2  & 7.6  & 28.2  & 56.1 & 38.2 & 38.2  & 9.2   & 29.3 & 25.9  & 21.6 & 28.3                                      \\
      \textbf{DDT(Ours)}               &                               &                                   & 56.1 & 66.6   & 39.4 & 55.2 & 51.3   & 79.9 & 62.1 & 8.3  & 57.9  & 46.3 & 40.5  & 39.3 & 51.4  & 96.3 & 84.5 & 60.9  & 28.0  & 42.7 & 56.7  & 59.2 & \textbf{55.6\footnotesize \textcolor{red}{+27.3}} \\ \bottomrule
    \end{tabular}%
  }
\end{table*}

%%%%%%%%%%%%%%%%%%%%%%%%%%%%%%%%%%%%%%%%%%%%%%%%%%%%%%%%%%%%%
\begin{table}[]
  \renewcommand{\arraystretch}{1.15}
  \setlength{\tabcolsep}{2pt}
   % 调整间距

  \centering
  \caption{Quantitative results on adaptation from VOC to Comic (V$\rightarrow$Co). The bold indicates the best results.}

  \label{tab:voc_to_comic}
  \resizebox{\columnwidth}{!}{%
    \begin{tabular}{ccc|cccccc|c}
      \toprule
      Method                                       & Reference                     & Detector                          & bicycle & bird & car  & cat  & dog  & person & mAP                                                \\ \midrule
      \textbf{DA-Faster} ~\cite{chen2018da-faster} & \small{\textit{CVPR'18}} & FRCNN\footnotesize -V16                   & 31.1    & 10.3 & 15.5 & 12.4 & 19.3 & 39.0   & 21.2                                               \\
      \textbf{SWDA} ~\cite{saito2019swda}          & \small{\textit{CVPR'19}} & FRCNN\footnotesize -V16                   & 36.4    & 21.8 & 29.8 & 15.1 & 23.5 & 49.6   & 29.4                                               \\
      \textbf{STABR} ~\cite{stabr}                 & \small{\textit{CVPR'19}} & SSD\footnotesize -V16                     & 50.6    & 13.6 & 31.0 & 7.5  & 16.4 & 41.4   & 26.8                                               \\
      \textbf{MCRA}   ~\cite{mcda}                 & \small{\textit{ECCV'20}} & FRCNN\footnotesize -V16                   & 47.9    & 20.5 & 37.4 & 20.6 & 24.5 & 50.2   & 33.5                                               \\
      \textbf{I3Net} ~\cite{i3net}                 & \small{\textit{CVPR'21}} & SSD\footnotesize -V16                     & 47.5    & 19.9 & 33.2 & 11.4 & 19.4 & 49.1   & 30.1                                               \\
      \textbf{DBGL}  ~\cite{dbgl}                  & \small{\textit{ICCV'21}} & FRCNN\footnotesize -R101                  & 35.6    & 20.3 & 33.9 & 16.4 & 26.6 & 45.3   & 29.7                                               \\
      \textbf{D-ADAPT} ~\cite{dadapt}              & \small{\textit{ICLR'22}} & FRCNN\footnotesize -R101                  & 52.4    & 25.4 & 42.3 & 43.7 & 25.7 & 53.5   & 40.5                                               \\ \midrule \midrule
      Baseline                                     & \multirow{2}{*}{/}            & \multirow{2}{*}{FRCNN\footnotesize -R18}  & 29.4    & 6.8  & 11.6 & 5.6  & 7.4  & 26.3   & 14.5                                               \\
      \textbf{DDT(Ours)}                           &                               &                                   & 56.8    & 21.2 & 46.6 & 25.5 & 32.3 & 72.6   & 42.5\footnotesize \textcolor{red}{\textbf{+28.0}}          \\
      Baseline                                     & \multirow{2}{*}{/}            & \multirow{2}{*}{FRCNN\footnotesize -R50}  & 37.1    & 6.9  & 29.9 & 6.9  & 10.5 & 30.0   & 20.2                                               \\
      \textbf{DDT(Ours)}                           &                               &                                   & 60.9    & 28.2 & 52.5 & 29.3 & 36.8 & 74.5   & 47.0\footnotesize \textcolor{red}{\textbf{+26.8}}          \\
      Baseline                                     & \multirow{2}{*}{/}            & \multirow{2}{*}{FRCNN\footnotesize -R101} & 37.0    & 6.8  & 31.2 & 4.8  & 7.2  & 26.8   & 19.0                                               \\
      \textbf{DDT(Ours)}                           &                               &                                   & 63.2    & 34.8 & 56.6 & 31.7 & 39.0 & 75.9   & \textbf{50.2}\footnotesize \textcolor{red}{\textbf{+31.2}} \\ \bottomrule
    \end{tabular}%
  }
\end{table}

%%%%%%%%%%%%%%%%%%%%%%%%%%%%%%%%%%%%%%%%%%%%%%%%%%%%%%%%%%%%%%%%%%%

\begin{table}[]
  \renewcommand{\arraystretch}{1.15}
  \setlength{\tabcolsep}{3pt}
   % 调整间距

  \centering
  \caption{Quantitative results on adaptation from VOC to Watercolor (V$\rightarrow$W). The bold indicates the best results.}

  \label{tab:voc_to_watercolor}
  \resizebox{\columnwidth}{!}{%
    \begin{tabular}{ccc|cccccc|c}
      \toprule
      Method                                   & Reference                      & Detector                          & bicycle & bird & car  & cat  & dog  & person & mAP                                                \\ \midrule
      \textbf{SWDA}  ~\cite{chen2018da-faster} & \small{\textit{CVPR‘19}}  & FRCNN\footnotesize -V16                   & 82.3    & 55.9 & 46.5 & 32.7 & 35.5 & 66.7   & 53.3                                               \\
      \textbf{MCRA} ~\cite{mcra}               & \small{\textit{ECCV‘20}}  & FRCNN\footnotesize -V16                   & 87.9    & 52.1 & 51.8 & 41.6 & 33.8 & 68.8   & 56.0                                               \\
      \textbf{UMT}  ~\cite{umt}                & \small{\textit{CVPR’21}}  & FRCNN\footnotesize -R101                  & 88.2    & 55.3 & 51.7 & 39.8 & 43.6 & 69.9   & 58.1                                               \\
      \textbf{IIOD}  ~\cite{iiod}              & \small{\textit{TPAMI‘21}} & FRCNN\footnotesize -V16                   & 95.8    & 54.3 & 48.3 & 42.4 & 35.1 & 65.8   & 56.9                                               \\
      \textbf{I3Net}  ~\cite{i3net}            & \small{\textit{CVPR’21}}  & SSD\footnotesize -V16                     & 81.1    & 49.3 & 46.2 & 35.0 & 31.9 & 65.7   & 51.5                                               \\
      \textbf{SADA}  ~\cite{sada}              & \small{\textit{IJCV‘21}}  & FRCNN\footnotesize -R50                   & 82.9    & 54.6 & 52.3 & 40.5 & 37.7 & 68.2   & 56.0                                               \\
      \textbf{CDG}  ~\cite{cdg}                & \small{\textit{AAAI’21}}  & FRCNN\footnotesize -V16                   & 97.7    & 53.1 & 52.1 & 47.3 & 38.7 & 68.9   & 59.7                                               \\
      \textbf{VDD}   ~\cite{vdd}               & \small{\textit{ICCV‘21}}  & FRCNN\footnotesize -V16                   & 90.0    & 56.6 & 49.2 & 39.5 & 38.8 & 65.3   & 56.6                                               \\
      \textbf{DBGL}   ~\cite{dbgl}             & \small{\textit{ICCV’21}}  & FRCNN\footnotesize -R101                  & 83.1    & 49.3 & 50.6 & 39.8 & 38.7 & 61.3   & 53.8                                               \\
      \textbf{AT}  ~\cite{li2022AT}            & \small{\textit{CVPR’22}}  & FRCNN\footnotesize -V16                   & 93.6    & 56.1 & 58.9 & 37.3 & 39.6 & 73.8   & 59.9                                               \\
      \textbf{LODS}  ~\cite{lods}              & \small{\textit{CVPR’22}}  & FRCNN\footnotesize -R101                  & 95.2    & 53.1 & 46.9 & 37.2 & 47.6 & 69.3   & 58.2                                               \\ \midrule \midrule
      Baseline                                 & \multirow{2}{*}{/}             & \multirow{2}{*}{FRCNN\footnotesize -R18}  & 71.4    & 36.4 & 39.1 & 19.9 & 12.7 & 52.0   & 38.6                                               \\
      \textbf{DDT(Ours)}                       &                                &                                   & 81.9    & 53.7 & 54.3 & 37.7 & 31.5 & 68.3   & 54.6\footnotesize \textcolor{red}{\textbf{+16.0}}          \\
      Baseline                                 & \multirow{2}{*}{/}             & \multirow{2}{*}{FRCNN\footnotesize -R50}  & 68.2    & 40.2 & 44.7 & 21.7 & 10.3 & 44.0   & 38.2                                               \\
      \textbf{DDT(Ours)}                       &                                &                                   & 96.4    & 58.6 & 52.6 & 33.7 & 36.2 & 71.9   & 58.2\footnotesize \textcolor{red}{\textbf{+20.0}}          \\
      Baseline                                 & \multirow{2}{*}{/}             & \multirow{2}{*}{FRCNN\footnotesize -R101} & 72.5    & 40.1 & 45.7 & 30.5 & 18.1 & 45.8   & 42.1                                               \\
      \textbf{DDT(Ours)}                       &                                &                                   & 87.1    & 64.0 & 55.7 & 50.6 & 48.8 & 75.7   & \textbf{63.7}\footnotesize \textcolor{red}{\textbf{+21.6}} \\ \bottomrule
    \end{tabular}%
  }
\end{table}

\textbf{Real to Artistic adaptation.} In Tab. ~\ref{tab:voc_to_clipart}, ~\ref{tab:voc_to_comic}, and ~\ref{tab:voc_to_watercolor}, we show the results of real to artistic cross-domain object detection.
Our results of Resnet50 and ResNet101 significantly surpass the previous best method AT ~\cite{li2022AT} by 3.9 and 4.8 mAP, respectively.
On Comic ~\cite{clipart_comic_watercolor}, the results of ResNet18, ResNet50, and ResNet101 ~\cite{resnet} greatly surpass the previous best result of AT ~\cite{li2022AT}, by 2.0, 6.5, and 9.7 mAP, respectively. Similarly, our best result surpasses AT ~\cite{li2022AT} by 3.8 mAP on Watercolor ~\cite{clipart_comic_watercolor} as shown in Tab. ~\ref{tab:voc_to_watercolor}.
In real to artistic benchmark, overall, we find that due to the significant differences between real-world images and artistic-style images, the cross-domain performance is poor. Compared to the baseline, our method exhibits an average relative improvement of 95\%, 163\%, and 48\% on Clipart, Comic, and Watercolor, respectively. This indicates that real to artistic adaptation is a challenging task, and it also demonstrates that our approach we have successfully improved the cross-domain performance by reducing the gap between real and artistic domain.

\begin{table}[]
  \renewcommand{\arraystretch}{1.15}
  \setlength{\tabcolsep}{2pt}
   % 调整间距
  
  \centering
  \caption{Results of our diffusion feature extractor (Diff.) compared to other backbones. The bold and underlined represent the best and second performances, respectively}
  \label{tab:backbone experiments}
  \resizebox{\columnwidth}{!}{%
  \begin{tabular}{ccccc|cccc|ccc}
  \toprule
              Backbones& \rotatebox{45}{\small \textbf{V}$\rightarrow$\textbf{V}} & \rotatebox{45}{\small \textbf{Ca}$\rightarrow$\textbf{Ca}} & \rotatebox{45}{\small \textbf{V}$\rightarrow$\textbf{Ca}} & Rel.(\%)    & \rotatebox{45}{\small \textbf{S}$\rightarrow$\textbf{S}} & \rotatebox{45}{\small \textbf{C\footnotesize{S}}$\rightarrow$\textbf{C\footnotesize{S}}} & \rotatebox{45}{\small \textbf{S}$\rightarrow$\textbf{C\footnotesize{S}}} & Rel.(\%)   & \small \rotatebox{45}{\textbf{B}$\rightarrow$\textbf{B}} & \small \rotatebox{45}{\textbf{S}$\rightarrow$\textbf{B}} & Rel.(\%)   \\ \midrule
  R101  \small ~\cite{resnet}         & 84.0   & 40.0   & 28.3   & 70.8  & 83.0   & 72.6   & 43.4   & 59.8 & 74.2   & 34.2   & 46.1 \\
  ConvNext\small{-Base} ~\cite{Convnext}    & \textbf{91.5}   & \textbf{62.8}   & \underline{44.1}   & 70.2  & \textbf{87.9}   & \textbf{78.2}   & \textbf{61.8}   & \underline{79.0} & \textbf{80.5}   & 44.8   & 55.6 \\
  Swin\small{-Base} ~\cite{swin}       & 86.9   & 51.6   & 32.2   & 62.4  & \underline{87.6}   & 73.7   & 53.7   & 72.8 & 79.7   & 39.3   & 49.4 \\
  VIT\small{-Base} ~\cite{vit}           & 84.9   & 31.5   & 28.6   & \underline{90.8}  & 77.6   & 72.1   & 48.4   & 67.1 & 73.4   & 40.9   & 55.7 \\
  MAE \small{(VIT-Base)} ~\cite{mae,vit}         & 85.8   & 37.8   & 26.4   & 69.8  & 85.0   & \underline{77.7}   & 57.9   & 74.5 & 78.3   & 40.9   & 52.2 \\
  GLIP \small{(Swin-Tiny)} ~\cite{glip,swin}        & \underline{88.1}   & \underline{55.6}   & 39.8   & 71.6  & 86.5   & 77.2   & \underline{59.3}   & 76.8 & \underline{79.7}   & \textbf{50.4}   & \underline{63.3} \\
  Diff (Ours) & 75.4   & 43.9   & \textbf{47.4}   & \textbf{108.1} & 76.1   & 71.7   & 58.2   & \textbf{81.2} & 71.8   & \underline{50.1}   & \textbf{69.8} \\ \bottomrule
  \end{tabular}%
  }
  \end{table}

% \usepackage{graphicx}
\begin{table}[]
  \renewcommand{\arraystretch}{1.1}
  \setlength{\tabcolsep}{4pt}
   % 调整间距

  \centering
  \caption{Results of using our diffusion backbone (Diff.) as the teacher model to train student models on different backbones. The bold indicates the best results.}
  \label{tab:teacher experiments}
  \resizebox{\columnwidth}{!}{%
  
    \begin{tabular}{cc|ccc}
      \toprule
      \multicolumn{2}{c}{Detector Setting}     & \multicolumn{3}{c}{Cross Domain Settings}                                                                                                                                                                \\ \midrule
      Student                                  & Teacher                                   & \textbf{V}$\rightarrow$\textbf{Ca}                 & \textbf{S}$\rightarrow$\textbf{Cs}                 & \textbf{S}$\rightarrow$\textbf{B}                  \\ \midrule
      R101 \footnotesize ~\cite{resnet}                & ConvNext\footnotesize -Base ~\cite{Convnext}      & 43.2\footnotesize \textcolor{red}{+19.9}                   & 60.2\footnotesize \textcolor{red}{+16.8}                   & 54.6\footnotesize \textcolor{red}{+20.4}                   \\
      R101 \footnotesize ~\cite{resnet}                & Swin\footnotesize -Base ~\cite{swin}              & 35.1\footnotesize \textcolor{red}{+6.8}                    & 57.5\footnotesize \textcolor{red}{+14.1}                   & 52.9\footnotesize \textcolor{red}{+18.7}                   \\
      R101 \footnotesize ~\cite{resnet}                & VIT\footnotesize -Base ~\cite{vit}                & 37.4\footnotesize \textcolor{red}{+9.1}                    & 54.8\footnotesize \textcolor{red}{+11.4}                   & 51.1\footnotesize \textcolor{red}{+16.9}                   \\
      R101  \footnotesize ~\cite{resnet}               & MAE \footnotesize (VIT-Base) ~\cite{mae,vit}      & 35.8\footnotesize \textcolor{red}{+7.5}                    & 59.0\footnotesize \textcolor{red}{+16.2}                   & 53.2\footnotesize \textcolor{red}{+19.0}                   \\
      R101 \footnotesize ~\cite{resnet}                & GLIP \footnotesize (Swin-Tiny) ~\cite{glip,swin}  & 39.6\footnotesize \textcolor{red}{+11.3}                   & 58.9\footnotesize \textcolor{red}{+15.5}                   & 54.0\footnotesize \textcolor{red}{+19.8}                   \\
      R101 \footnotesize ~\cite{resnet}                & Diff. (Ours)                              & \textbf{55.6}\footnotesize \textcolor{red}{\textbf{+27.3}} & \textbf{64.0}\footnotesize \textcolor{red}{\textbf{+20.6}} & \textbf{58.3}\footnotesize \textcolor{red}{\textbf{+24.1}} \\ \midrule

      ConvNext\footnotesize -Base  ~\cite{Convnext}    & Diff. (Ours)                              & \textbf{59.5}\footnotesize \textcolor{red}{\textbf{+18.4}} & 63.5\footnotesize \textcolor{red}{+1.7}                    & \textbf{59.6}\footnotesize \textcolor{red}{\textbf{+7.1}}  \\
      Swin\footnotesize -Base  ~\cite{swin}            & Diff. (Ours)                              & 46.6\footnotesize \textcolor{red}{+14.4}                   & 63.6\footnotesize \textcolor{red}{+9.9}                    & 58.6\footnotesize \textcolor{red}{+13.8}                   \\
      VIT\footnotesize -Base  ~\cite{vit}              & Diff. (Ours)                              & 41.7\footnotesize \textcolor{red}{+13.1}                   & 60.2\footnotesize \textcolor{red}{+11.8}                   & 54.9\footnotesize \textcolor{red}{+15.6}                   \\
      MAE \footnotesize (VIT-Base) ~\cite{mae,vit}     & Diff. (Ours)                              & 43.1\footnotesize \textcolor{red}{+16.7}                   & \textbf{64.6}\footnotesize \textcolor{red}{\textbf{+6.7}}  & 57.3\footnotesize \textcolor{red}{+16.4}                   \\
      GLIP \footnotesize (Swin-Tiny) ~\cite{glip,swin} & Diff. (Ours)                              & 49.5\footnotesize \textcolor{red}{+9.7}                    & 64.0\footnotesize \textcolor{red}{+4.7}                    & 59.5\footnotesize \textcolor{red}{+9.4}                    \\ \bottomrule
    \end{tabular}%
  }
\end{table}

\begin{table}[]
   % 调整间距
  \renewcommand{\arraystretch}{1.15}
  \setlength{\tabcolsep}{2pt}
  \centering
  \caption{Results of ablation experiments on Diffusion Teacher and Mean Teacher.}
  \label{tab:duffsion_mean_teacher}
  \resizebox{\columnwidth}{!}{%
  
    \begin{tabular}{l|cccccc}
      \toprule
      Settings of Self-training
                                     & \textbf{Cs}$\rightarrow$\textbf{B}   & \textbf{S}$\rightarrow$\textbf{Cs}   & \textbf{S}$\rightarrow$\textbf{B}    & \textbf{V}$\rightarrow$\textbf{Ca}    & \textbf{V}$\rightarrow$\textbf{Co}    & \textbf{V}$\rightarrow$\textbf{W}    \\ \midrule
      DDT (R101)                     & 43.4                                 & 64.0                                 & 58.3                                 & 55.6                                  & 50.2                                  & 63.7                                 \\
      \textbf{w/o} Mean Teacher      & 40.6\footnotesize \color[HTML]{009901}{-2.8} & 62.5\footnotesize \color[HTML]{009901}{-1.5} & 55.5\footnotesize \color[HTML]{009901}{-2.8} & 49.1\footnotesize \color[HTML]{009901}{-6.5}  & 46.2\footnotesize \color[HTML]{009901}{-4.0}  & 61.0\footnotesize \color[HTML]{009901}{-2.7} \\
      \textbf{w/o} Diffusion Teacher & 40.1\footnotesize \color[HTML]{009901}{-3.3} & 57.4\footnotesize \color[HTML]{009901}{-6.6} & 54.4\footnotesize \color[HTML]{009901}{-3.9} & 46.9\footnotesize \color[HTML]{009901}{-8.7}  & 37.5\footnotesize \color[HTML]{009901}{-12.7} & 57.4\footnotesize \color[HTML]{009901}{-6.3} \\
      \textbf{w/o} All Teacher       & 37.5\footnotesize \color[HTML]{009901}{-5.9} & 56.1\footnotesize \color[HTML]{009901}{-7.9} & 53.7\footnotesize \color[HTML]{009901}{-4.6} & 41.9\footnotesize \color[HTML]{009901}{-13.7} & 37.4\footnotesize \color[HTML]{009901}{-12.8} & 57.1\footnotesize \color[HTML]{009901}{-6.6} \\ \bottomrule
    \end{tabular}%
  }
\end{table}

% \usepackage{graphicx}
\begin{table}[]
  \renewcommand{\arraystretch}{1.15}
  \setlength{\tabcolsep}{1pt}
   % 调整间距
  \centering
  \caption{Ablation results of the diffusion backbone under different time steps and save steps.}
  \label{tab:diffusion setting}
  \small
  \resizebox{\columnwidth}{!}{%
    \begin{tabular}{cc|cc|cccccc}
      \toprule
      \begin{tabular}[c]{@{}c@{}}Time \\ Steps\end{tabular} & \begin{tabular}[c]{@{}c@{}}Save \\ Steps\end{tabular} & \begin{tabular}[c]{@{}c@{}}Train Time\\  (s/iter)\end{tabular} & \begin{tabular}[c]{@{}c@{}}Inf. Time\\  (ms/image)\end{tabular} & \textbf{Cs}$\rightarrow$\textbf{B} & \textbf{S}$\rightarrow$\textbf{Cs} & \textbf{S}$\rightarrow$\textbf{B} & \textbf{V}$\rightarrow$\textbf{Ca} & \textbf{V}$\rightarrow$\textbf{Co} & \textbf{V}$\rightarrow$\textbf{W} \\ \midrule
      1                              & 1                              & 0.82                           & 271.1                          & 29.8                               & 57.2                               & 50.4                              & 37.8                               & 36.9                               & 52.7                              \\
      2                              & 2                              & 1.14                           & 402.5                          & 31.0                               & 57.5                               & 50.1                              & 39.3                               & 36.3                               & 51.8                              \\
      \underline{\textbf{5}}         & \underline{\textbf{5}}         & 1.56                           & 780.4                          & \textbf{32.7}                               & \textbf{58.2}                               & 50.1                              & 47.4                               & 39.4                               & 53.8                              \\
      10                             & 5                              & 2.84                           & 1424.2                         & 29.8                               & 56.3                               & \textbf{51.1}                              & 48.4                               & 40.9                               & 53.7                              \\
      20                             & 10                             & 5.48                           & 2710.2                         & 28.0                               & 55.2                               & 50.2                              & \textbf{48.8}                               & \textbf{41.3}                               & \textbf{54.6}                              \\ \bottomrule
    \end{tabular}%
  }

\end{table}

\begin{figure}
   % 调整间距
  
  \centering
  \begin{minipage}[b]{0.45\textwidth}
    \subfigure[Confusion matrix of \textbf{baseline} (left) and \textbf{DDT} (right)]{
      \includegraphics[width=0.5\textwidth]{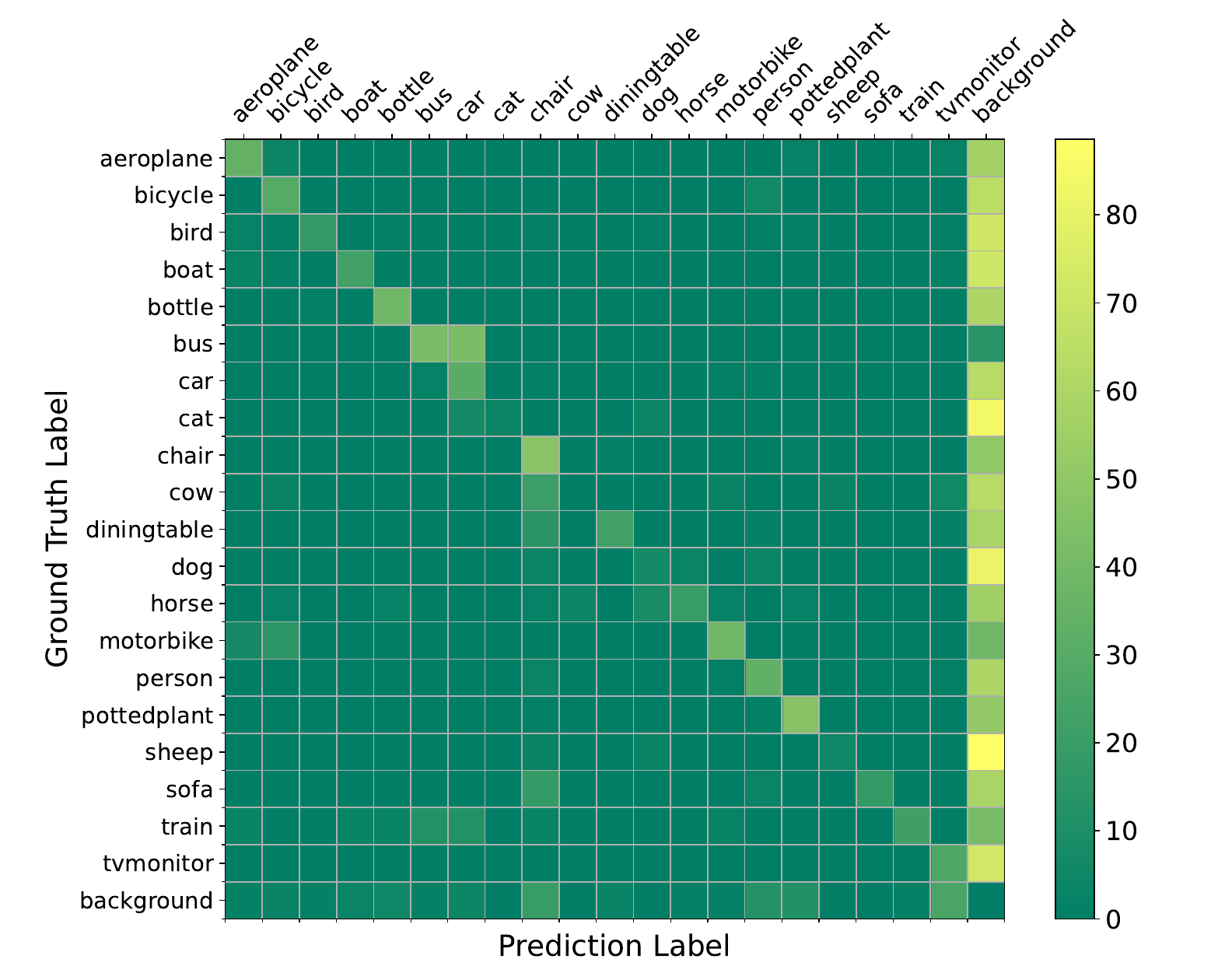}
      \includegraphics[width=0.5\textwidth]{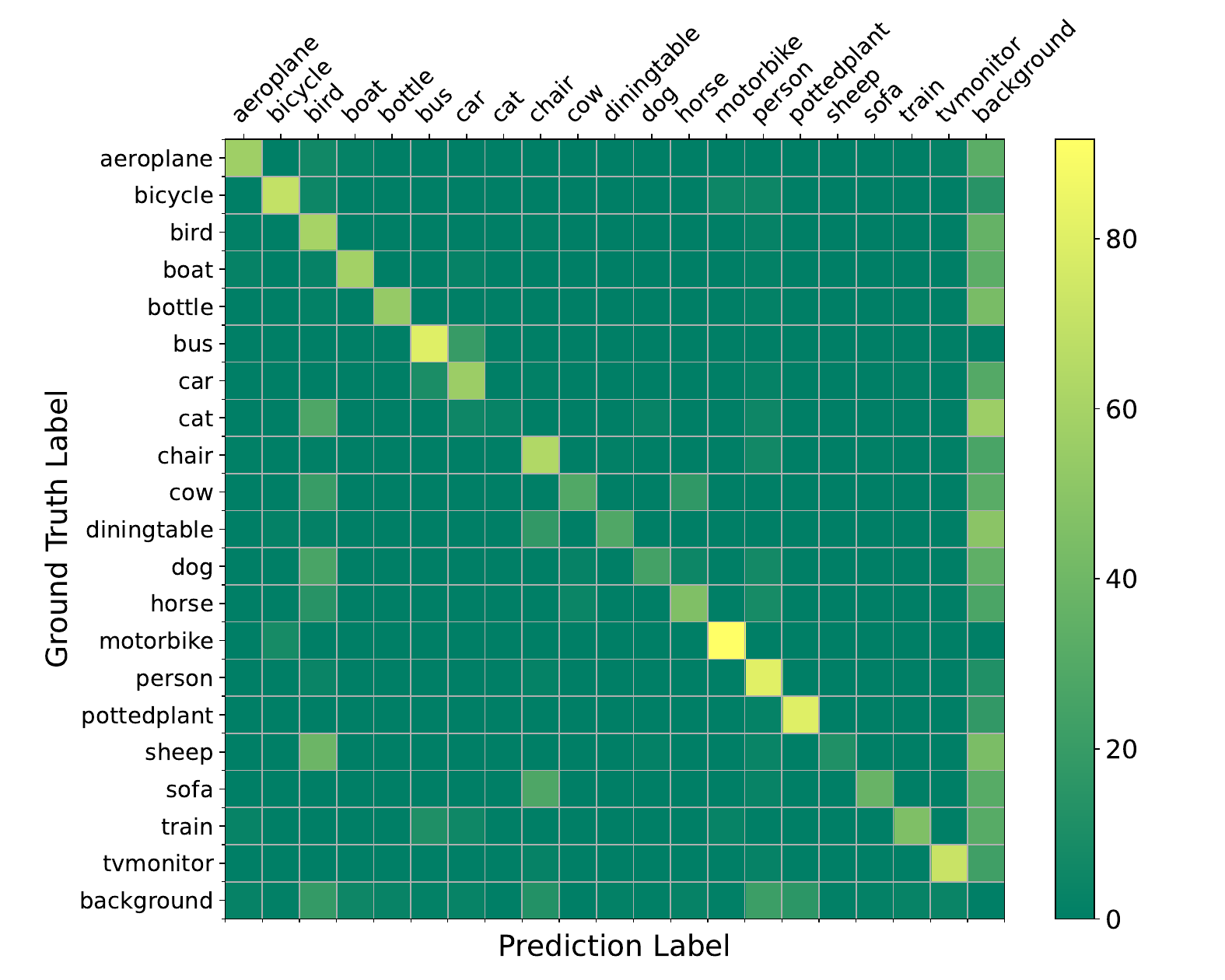}
      \label{fig:confuse matrix}
    }
  \end{minipage}

  \begin{minipage}[b]{0.45\textwidth}
    \subfigure[COCO ~\cite{coco} style detection error analysis of the \textbf{baseline} (left) and \textbf{DDT} (right)]{
      \includegraphics[width=0.5\textwidth]{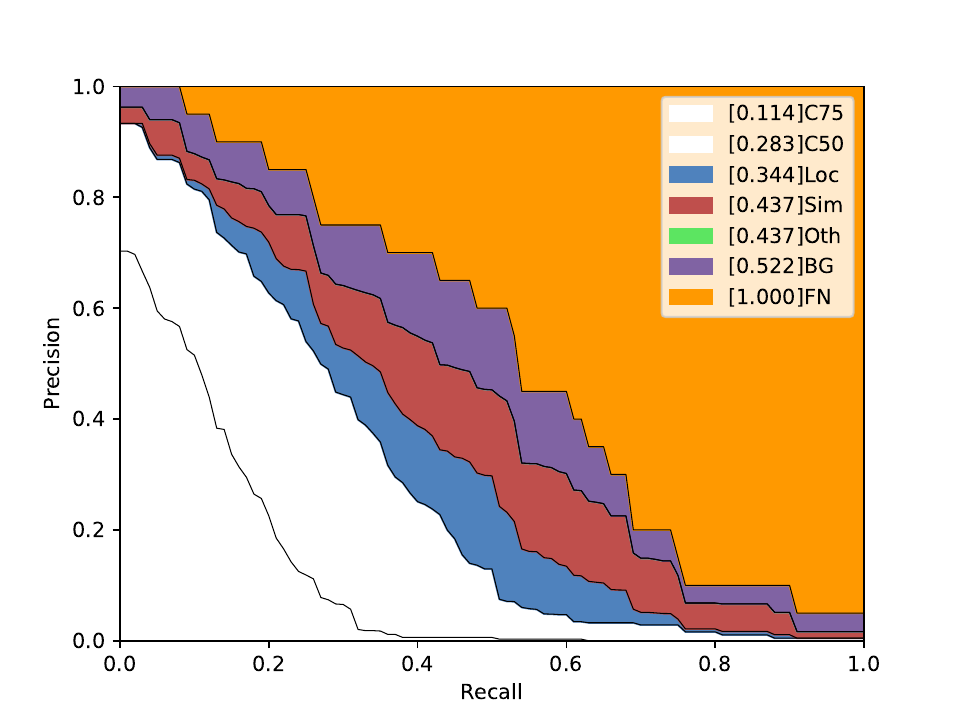}
      \includegraphics[width=0.5\textwidth]{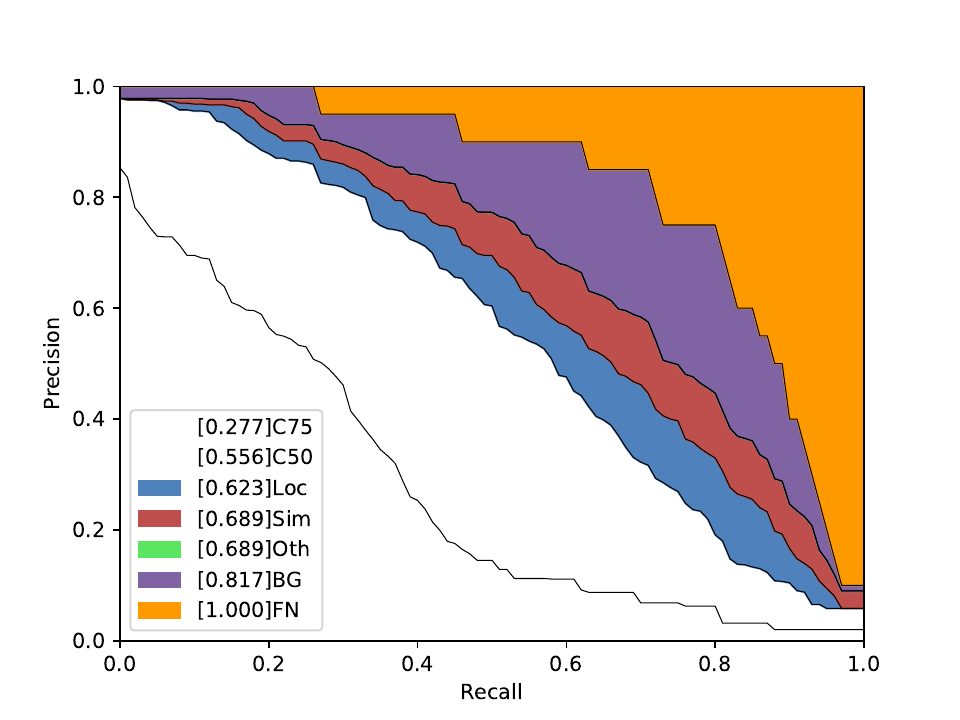}
      \label{fig:coco error}
    }
  \end{minipage}
  \caption{Error analysis on Clipart. \textnormal{It is evident that our method significantly reduces false negatives, which correspond to missed detections.}}
  \label{fig:error analysis}
\end{figure}

\subsection{Ablation Studies}

We conduct additional experiments to analyze the feature representation capabilities of different models. Specifically, we compared our diffusion model with powerful backbones, including ConvNext ~\cite{Convnext}, Swin Transformer ~\cite{swin}, VIT ~\cite{vit}, as well as the self-supervised method MAE ~\cite{mae}, pretrained on ImageNet \cite{imagenet}. Additionally, GLIP ~\cite{glip}, which is pretrained on a larger dataset and has shown promising performance on object detection benchmarks. Our objective is to investigate two questions:
\begin{enumerate}
  \item Will the diffusion model offer better intra-domain and cross-domain feature representation?
  \item Will the diffusion model serve as a better teacher?
\end{enumerate}

\textbf{Ablation Study on the Intra-domain and Cross-domain Representation.} To answer the first question, we evaluate the performance of seven models across different data settings in Tab. ~\ref{tab:backbone experiments}. Specifically, we compared their intra-domain performance in training and testing within the source domain (\textbf{V}$\rightarrow$\textbf{V}, \textbf{S}$\rightarrow$\textbf{S}) and target domain (\textbf{Ca}$\rightarrow$\textbf{Ca}, \textbf{Cs}$\rightarrow$\textbf{Cs}, \textbf{B}$\rightarrow$\textbf{B}), and cross-domain testing (\textbf{V}$\rightarrow$\textbf{Ca}, \textbf{S}$\rightarrow$\textbf{Cs}, \textbf{S}$\rightarrow$\textbf{B}) to assess their cross-domain feature representation capabilities. we find that our diffusion model performe poorly within the intra-domain, lagging behind the other six models. In cross-domain testing, our diffusion model outperformed other methods on Clipart ~\cite{clipart_comic_watercolor} but remained inferior to ConvNext ~\cite{Convnext} and GLIP ~\cite{glip}. Additionally, we calculate the cross-domain metrics for each model and the results obtained from training and testing in the target domain to measure the relative cross-domain capabilities of each models, represented as "Rel." in Tab. \ref{tab:backbone experiments}. We find that the diffusion model consistently achieves the best relative cross-domain performance. Overall, the answer to the first question may be disappointing, as the diffusion detector showes some improvement in cross-domain performance but still fall behind the detectors with stronger and large-dataset pretrained backbones.

\textbf{Ablation Study of Different Teachers.} In Tab. ~\ref{tab:teacher experiments}, we present the performance of different teacher and student settings for cross-domain detection. First, we use ResNet101 ~\cite{resnet} as the student and other models as teacher. We find that although the diffusion model performs worse than ConvNext ~\cite{Convnext} and GLIP ~\cite{glip} in cross-domain performance, it exhibits significantly better performance when used as a teacher model. Furthermore, when we use the diffusion model as the teacher and the other six models as students, it consistently brings large improvements. This answers our second question, confirming that our diffusion model is indeed a better teacher, even when faced with highly competent students and consistently improve their performance.

\textbf{Ablation Results on Diffusion and Mean Teacher.} To better understand the significance of teachers in our DDT, we present the results of different teacher model settings in Tab. ~\ref{tab:duffsion_mean_teacher}. The findings reveal that excluding the Mean Teacher and Diffusion Teacher from our method leads to an average decrease of 3.1 and 6.7 mAP, respectively. When all teachers are removed, the self-training performance experiences an average decline of 8.3 mAP. These results clearly demonstrate that both the diffusion teacher and mean teacher play crucial roles in our DDT and are indispensable for achieving better performance. Fig. ~\ref{fig:training_process} provides an intuitive illustration of the impacts of the diffusion teacher and mean teacher in training process.

\textbf{Ablation Results of Different Diffusion Settings.} We report the results of the diffusion models with different time steps and save steps settings in Tab. ~\ref{tab:diffusion setting}. It is observed that in cross-domain detection with a larger domain gap (real to artistic), longer time steps and save steps show better results. We consider a trade-off between accuracy and efficiency and choose time steps 5 and save steps 5 as our default settings.

\subsection{Analysis}

%%%%%%%%%%%%%%%%%%%%%%%%%%%%%%%%%%%%%%%%%%%%%%%%%%%%%%%%%%%%%%%%%%%%%%%%%%%%%%%%%%%%%%%%%%%%%%%
% \begin{figure}[]
%   \centering
%   \begin{minipage}[b]{0.5\textwidth}
%     \includegraphics[width=0.3\textwidth]{figure/umap_images.pdf}
%     \includegraphics[width=0.3\textwidth]{figure/umap_features_r101.pdf}
%     \includegraphics[width=0.3\textwidth]{figure/umap_features_ddt.pdf}
%   \end{minipage}
%   \caption{UMAP visualization of images and features from VOC (green) and Clipart (orange). \textnormal{We observe significant distribution disparities between images \textbf{(left)} from VOC ~\cite{voc} and Clipart ~\cite{clipart_comic_watercolor} and features extracted from ResNet101 ~\cite{resnet} \textbf{(middle)}, indicating a large domain gap. However, the feature extracted by our DDT \textbf{(right)} from VOC and Clipart preform a closer distribution, indicating that our method reduces the domain discrepancy of feature and improves the performance of cross-domain detection.}}
%   \label{fig:umap}
% \end{figure}

%%%%%%%%%%%%%%%%%%%%%%%%%%%%%%%%%%%%%%%%%%%%%%%%%%%%%%%%%%%%%%%%%%%%%%%%%%%%%%%%%%%%%%%%%%%%%%%

\textbf{Analysis of Feature Representation of Diffusion Model.} The results in Tab. \ref{tab:backbone experiments} and \ref{tab:teacher experiments} further deepen our understanding of the feature representation of diffusion model and its advantages in cross-domain detection. In our view, the observed results can be attributed as: fully frozen weight and adaptation of the diffusion model with the light structure that aligns with the hierarchical feature outputs, limit its performance within the intro-domain compared to fully trainable models. However, when applied as a teacher, the diffusion model guides the student to achieve superior performance in cross-domain, surpassing even the teacher itself. We think that the improved cross-domain representation ability can be attributed to the inherent characteristics of the diffusion model as well as the advantages gained from supervised learning on the source domain. In contrast, other fully trainable teacher models often concentrate primarily on supervised learning on the source domain, resulting in homogeneous optimization and limited guidance for the student. As a result, it becomes challenging to enhance the performance of the students to the level achieved by the homogeneous teacher. These results provide compelling evidence for the advantages of the diffusion model in addressing cross-domain detection tasks.

\textbf{Error analysis.} Error analysis on Clipart ~\cite{clipart_comic_watercolor} reveals that false negatives, i.e., missed detections, are the main factor impacting the performance on the target domain as shown in Fig. ~\ref{fig:error analysis}. Our method significantly reduces the number of missed detections, thereby greatly improving the performance of cross-domain detection. A representation of prediction results and feature visualization further corroborate this conclusion, as depicted in the Fig. ~\ref{fig:prediction result}.

\section{Conclusion}

In this paper, we propose a domain adaptive method based on the diffusion model to address the performance degradation caused by the large gap between the source and target domain. We employ a frozen-weight diffusion model as the backbone and extract intermediate feature in the inversion process for the detection task, which we refer to as the diffusion teacher. Subsequently, we apply diffusion teacher in the self-training framework to generate pseudo labels on the unlabeled target domain, guiding the learning of the student model. Our method significantly improves cross-domain detection performance on six datasets, achieving an average improvement of 21.2\% mAP compared to the baseline, surpassing the current SOTA methods by an average of 5.7\% mAP, without compromising the inference speed. Furthermore, we validate the consistent performance improvement of our method in more extensive experiments for detectors with more powerful backbones, demonstrating the strong and universality domain adaptive capability of our approach.

\section{Acknowledgment}
The authors would like to thank Xiamen University and
Unmanned Aerial Vehicle (UAV) Laboratory for the funding and providing with all the necessary technical support.

\bibliographystyle{ACM-Reference-Format}
\bibliography{sample-base}

\clearpage
\appendix

\section{Appendix}

\begin{figure*}[t]
  % 调整间距
 \centering
 \includegraphics[width=1.0\linewidth]{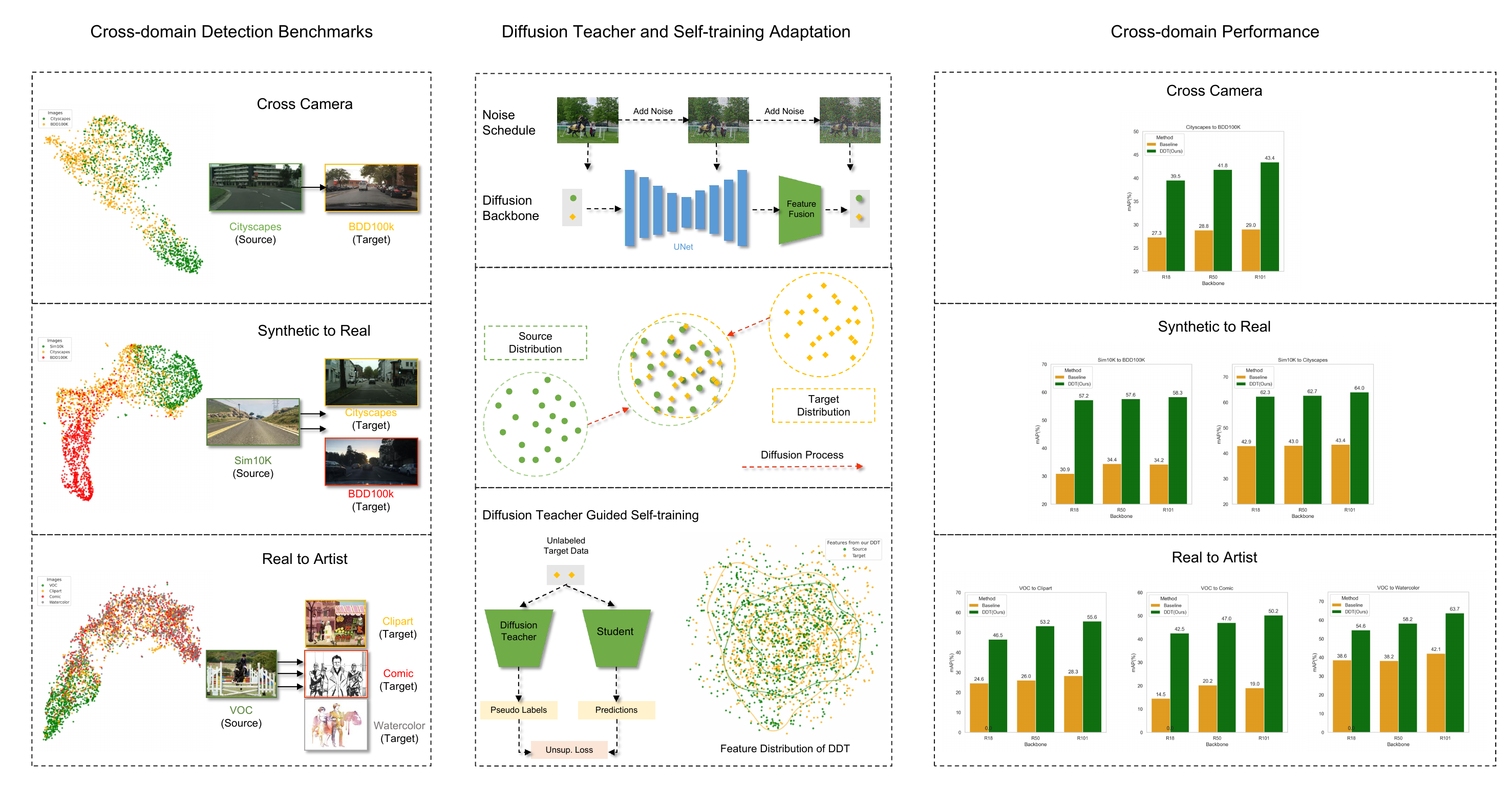}
 \caption{Main content of our work. 
 \textbf{Left}: We present three cross-domain detection benchmarks and visualize the image distributions from source and target domains using the UMAP method. It is evident that there is a large gap between different domains. 
 \textbf{Middle}: We utilize a frozen-weight diffusion model as the backbone to extract features, and employ a detector with the diffusion backbone as the teacher to guide the learning of the student model on the target domain with the self-training framework. Based on the visualization results of the feature distributions, our method significantly reduces the domain gap. 
 \textbf{Right}: Our method substantially enhances the performance of cross-domain object detection without increasing the inference cost. }
 \label{fig:introduction}
\end{figure*}

\subsection{Additional Ablation Studies}
In this section, we present additional ablation studies and comparisons to further investigate the effectiveness of our proposed Diffusion Domain Teacher (DDT).

\textbf{Ablation Study on Unsup Loss Weight $\lambda$.} We investigate the impact of different unsupervised loss weight $\lambda$ on cross-domain detection in Tab.\ref{tab:unsup weight}, including Cityscapes \cite{cityscapes} to BDD100K \cite{bdd100k} (\textbf{Cs}$\rightarrow$\textbf{B}), Sim10K \cite{sim10k} to Cityscapes \cite{cityscapes} (\textbf{S}$\rightarrow$\textbf{Cs}), and VOC \cite{voc} to Clipart \cite{clipart_comic_watercolor} (\textbf{V}$\rightarrow$\textbf{Ca}). Excessively high or low weight $\lambda$ will degrade the performance of cross-domain detection, so we simply set the parameter $\lambda$ to 1.

\textbf{Ablation Study on the Threshold  $\sigma$ of Pseudo Labels.} In Tab. \ref{tab:pseudo threshold}, we show the impact of different pseudo label threshold $\sigma$ on the final results. The threshold plays a crucial role in determining the quality of the generated pseudo labels during self-training, and inappropriate threshold will affect cross-domain performance. Ultimately, we choose 0.5 as the default threshold $\sigma$ setting in all our experiments.

\textbf{Ablation Study on Data Augmentation of Sup. and Unsup. Branches.} We follow previous work \cite{li2022AT} and utilize \textit{Strong Augmentation} and \textit{Weak Augmentation} during the self-training process. Tab. \ref{tab:data aug} presents the impact of \textit{Strong Augmentation} on both supervised and unsupervised data, highlighting the importance of data augmentation in the self-training process.

\textbf{Ablation Study on Different Versions of Stable Diffusion.} In Tab. \ref{tab:versions}, we present the results of using the popular Stable Diffusion V1.5 (SD-1.5) and the latest version (SD-2.1) respectively. Overall, the results of SD-1.5 are superior to those using SD-2.1, except for the cross-domain detection from Sim10K to BDD100K.

\textbf{Comparison of Backbone Efficiencies.} In Tab. ~\ref{tab:time experiments}, we present a comparative analysis of these models with respect to their architectural sizes, training cost, and inference latency. Diffusion model, with its substantial parameter count and protracted inference time, emerges as an impractical choice for deployment in routine detection tasks within operational settings. Nevertheless, the employment of this model as a strong teacher in self-training framework offers a strategic avenue to leverage its exceptional cross-domain prowess, achieving an average absolute improvement of 21.2 mAP and a relative improvement of 39.7\% compared to the baseline, without introducing any additional inference overhead.

% \usepackage{graphicx}
\begin{table}[h]
  \setlength{\tabcolsep}{10pt}
  \centering
  \caption{Ablation Result of Unsup. Loss Weight $\lambda$.}
  \label{tab:unsup weight}
 
  \begin{tabular}{c|ccc}
  \toprule
   $\lambda$ & \textbf{Cs}$\rightarrow$\textbf{B} & \textbf{S}$\rightarrow$\textbf{Cs} & \textbf{V}$\rightarrow$\textbf{Ca} \\ \midrule
  0.33                  & 42.7    & 63.5    & 54.7    \\
  0.50                  & 42.4    & 63.7    & 55.3    \\
  \underline{1.0}                   & \textbf{43.4}    & \textbf{64.0}    & \textbf{55.6}    \\
  2.0                   & 43.0    & \textbf{64.0}    & 53.3    \\
  3.0                   & 42.9    & 63.9    & 50.0    \\ \bottomrule
  \end{tabular}%
  \end{table}

% \usepackage{graphicx}
\begin{table}[]

  \setlength{\tabcolsep}{10pt}
  \centering
  \caption{Ablation Result of Threshold  $\sigma$ for Pseudo Labels.}
  \label{tab:pseudo threshold}

  \begin{tabular}{c|ccc}
  \toprule
  $\sigma$ & \textbf{Cs}$\rightarrow$\textbf{B} & \textbf{S}$\rightarrow$\textbf{Cs} & \textbf{V}$\rightarrow$\textbf{Ca} \\ \midrule
  0.3              & 41.9    & 63.7    & 52.4    \\
  0.4              & 42.3    & \textbf{64.2}    & 53.9    \\
  \underline{0.5}              & \textbf{43.4}    & 64.0    & 55.6    \\
  0.6              & 43.2    & 62.1    & \textbf{56.1}    \\
  0.7              & 43.0    & 60.3    & 53.2    \\ \bottomrule
  \end{tabular}%
  \end{table}

% \usepackage{graphicx}
\begin{table}[]
  \renewcommand{\arraystretch}{1.15}
  \setlength{\tabcolsep}{6pt}
  \centering
  \caption{Ablation Study of Data Augmentation.}
  \label{tab:data aug}

  \begin{tabular}{l|ccc}
  \toprule
  Settings of data Aug. & \textbf{Cs}$\rightarrow$\textbf{B} & \textbf{S}$\rightarrow$\textbf{Cs} & \textbf{V}$\rightarrow$\textbf{Ca} \\ \midrule
  DDT(Ours)             & 43.4    & 64.0    & 55.6    \\
  w/o Strong Aug. on Sup.    & 42.0\footnotesize \color[HTML]{009901}{-1.4}    & 62.0\footnotesize \color[HTML]{009901}{-2.0}    & 54.3\footnotesize \color[HTML]{009901}{-1.3}    \\
  w/o Strong Aug. on Unsup& 42.5\footnotesize \color[HTML]{009901}{-0.9}    & 62.2\footnotesize \color[HTML]{009901}{-1.8}    & 52.3\footnotesize \color[HTML]{009901}{-3.3}    \\
  w/o All Aug.     & 42.0\footnotesize \color[HTML]{009901}{-1.4}    & 61.8\footnotesize \color[HTML]{009901}{-2.2}    & 52.1\footnotesize \color[HTML]{009901}{-3.5}    \\ \bottomrule
  \end{tabular}%
  \end{table}

% \usepackage{multirow}
\begin{table}[]
  \renewcommand{\arraystretch}{1.15}
  \setlength{\tabcolsep}{2pt}
  \centering
  \caption{Ablation Result of Different Stable Diffusion Versions.}
  \label{tab:versions}
  \resizebox{\columnwidth}{!}{%
  \begin{tabular}{cccccccc}
  \toprule
  Detector                            & Version  & \textbf{Cs}$\rightarrow$\textbf{B}   & \textbf{S}$\rightarrow$\textbf{Cs}   & \textbf{S}$\rightarrow$\textbf{B}    & \textbf{V}$\rightarrow$\textbf{Ca}    & \textbf{V}$\rightarrow$\textbf{Co}    & \textbf{V}$\rightarrow$\textbf{W} \\ \midrule
  \multirow{2}{*}{Diffusion Detector} & \underline{SD-1.5}                      & 32.7    & 58.2    & 50.1   & \textbf{47.4}    & \textbf{44.4}    & \textbf{58.7}   \\
                                      & SD-2.1                      & \textbf{34.6}    & \textbf{58.9}    & \textbf{56.4}   & 45.4    & 42.8    & 55.2   \\ \midrule
  \multirow{2}{*}{DDT}                & \underline{SD-1.5}                      & \textbf{43.4}    & \textbf{64.0}      & 58.3   & \textbf{55.6}    & \textbf{50.2}    & \textbf{63.7}   \\
                                      & SD-2.1                      & 42.3    & 63.4    & \textbf{61.6}   & 53.7    & 48.9    & 63.3   \\ \bottomrule
  \end{tabular}
  }
  \end{table}

  \begin{table*}[b]
    \setlength{\tabcolsep}{6pt}
    \centering
    % \caption{\textnormal{Results of the efficiency comparison of detectors with different backbones applied a size of (1333, 800). Our DDT method significantly improves cross-domain detection performance across three different depths of backbone configurations, achieving an average absolute improvement of  \textbf{21.2} mAP and a relative improvement of \textbf{39.7}$\%$ compared to training solely on the source domain, without introducing any additional inference overhead.}}
    \caption{Results of the efficiency comparison of detectors with different backbones.}
    \label{tab:time experiments}
      \begin{tabular}{ccccccc}
        \toprule
        Methods                                  & Params (M) & Flops (G) & \begin{tabular}[c]{@{}c@{}}Train Time \\(s/iter)\end{tabular} & \begin{tabular}[c]{@{}c@{}}Inference Time\\  (ms/image)\end{tabular} & \begin{tabular}[c]{@{}c@{}}Average \\ Gain (mAP)\end{tabular} & \begin{tabular}[c]{@{}c@{}}Average Rel. \\ Improve (\%)\end{tabular} \\ \midrule
        ConvNext\footnotesize -Base ~\cite{Convnext}     & 105        & 401       & 0.672                          & 73.2                           & /                              & /                              \\
        Swin\footnotesize -Base  ~\cite{swin}            & 104        & 413       & 0.693                          & 79.4                           & /                              & /                              \\
        VIT\footnotesize -Base ~\cite{vit}               & 107        & 605       & 0.634                          & 116.5                          & /                              & /                              \\
        MAE \footnotesize (VIT-Base) ~\cite{mae,vit}     & 107        & 605       & 0.634                          & 116.5                          & /                              & /                              \\
        GLIP \footnotesize (Swin-Tiny) ~\cite{glip,swin} & 45         & 198       & 0.486                          & 84.1                           & /                              & /                              \\
        Diff. (Ours)                             & 991        & 8,256     & 1.56                           & 780.4                          & /                              & /                              \\ \midrule 
        DDT (R18)                                & 28         & 137       & 0.82                           & 24.6                           & \textbf{19.0}                  & \textbf{38.9}                  \\
        DDT (R50)                                & 41         & 184       & 0.90                           & 33.6                           & \textbf{21.7}                  & \textbf{40.1}                  \\
        DDT (R101)                               & 60         & 262       & 1.04                           & 40.9                           & \textbf{22.9}                  & \textbf{40.2}                  \\ \bottomrule
      \end{tabular}
  \end{table*}

\subsection{Additional Experiments}
In this section, we showcase further experiments, including: (1) Cityscapes \cite{cityscapes} to FoggyCityscapes \cite{foggy}, aiming to validate the results of adverse weather adaptation, and (2) the outcomes of our DDT method employing the FCOS \cite{tian2020fcos} detector, aiming to assess the performance of our approach across different detectors.

\textbf{Adverse Weather Adaptation from Cityscapes to FoggyCityscapes.} We present the result of adverse weather adaptation in Tab. \ref{tab:city_to_foggy}. Compared to the baseline, our method with R18, R50, and R101 show improvements of 11.8, 18.0, and 21.6 mAP, respectively. However, our method (50.0 mAP) still falls short of surpassing CMT \cite{cao2023cmt} (50.3 mAP) and HT \cite{deng2023HT} (50.4 mAP). FoggyCityscapes \cite{foggy} is a dataset where foggy weather conditions are added to Cityscapes, with similar images and identical labels from Cityscapes. We observe that our DDT method does not demonstrate superior performance in inter-domain as train and test on Cityscapes, which might be the reason for our weaker performance on the FoggyCityscapes compared to the current state-of-the-art results.

\textbf{Results of Adaptation with FCOS Detector.} Previous domain adaptation methods for detection primarily employ the Faster RCNN and FCOS detectors. In main text, we report the results of our method using the Faster RCNN \cite{chen2018da-faster} detector. To further validate the effectiveness of our approach, we also present the performance using the FCOS \cite{tian2020fcos} detector in Tab. \ref{tab:city_to_bdd100k_fcos}, \ref{tab:sim10k_to_bdd100k_fcos}, \ref{tab:sim10k_to_city_fcos}, \ref{tab:voc_to_comic_fcos}, \ref{tab:voc_to_watercolor_fcos}, \ref{tab:voc_to_clipart_fcos}. Overall, the FCOS detector yielded results comparable to the Faster RCNN and outperforms the previous SOTA results in five out of the six datasets, with the exception of Sim10K to Cityscapes where it falls short of HT \cite{deng2023HT}.

\subsection{Additional visualization results}

\begin{figure*}[t]
  % 调整间距
 \centering
 \includegraphics[width=1.0\linewidth]{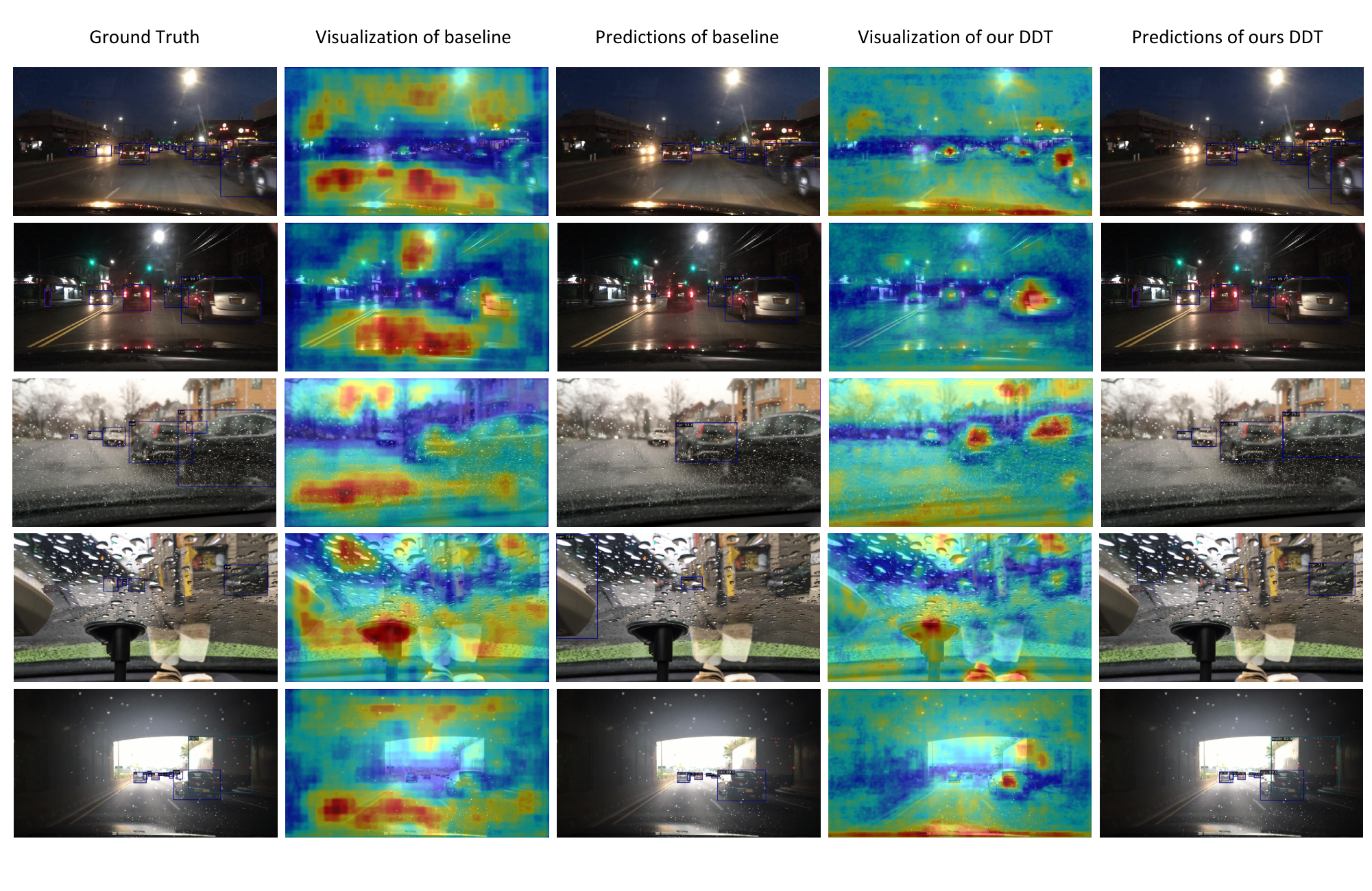}
 \caption{Qualitative prediction results and feature visualization of baseline and our DDT from Cityscapes to BDD100K.}
 \label{fig:bdd100k_for_cityscapes}
\end{figure*}

\begin{figure*}[t]
  % 调整间距
 \centering
 \includegraphics[width=0.9\linewidth]{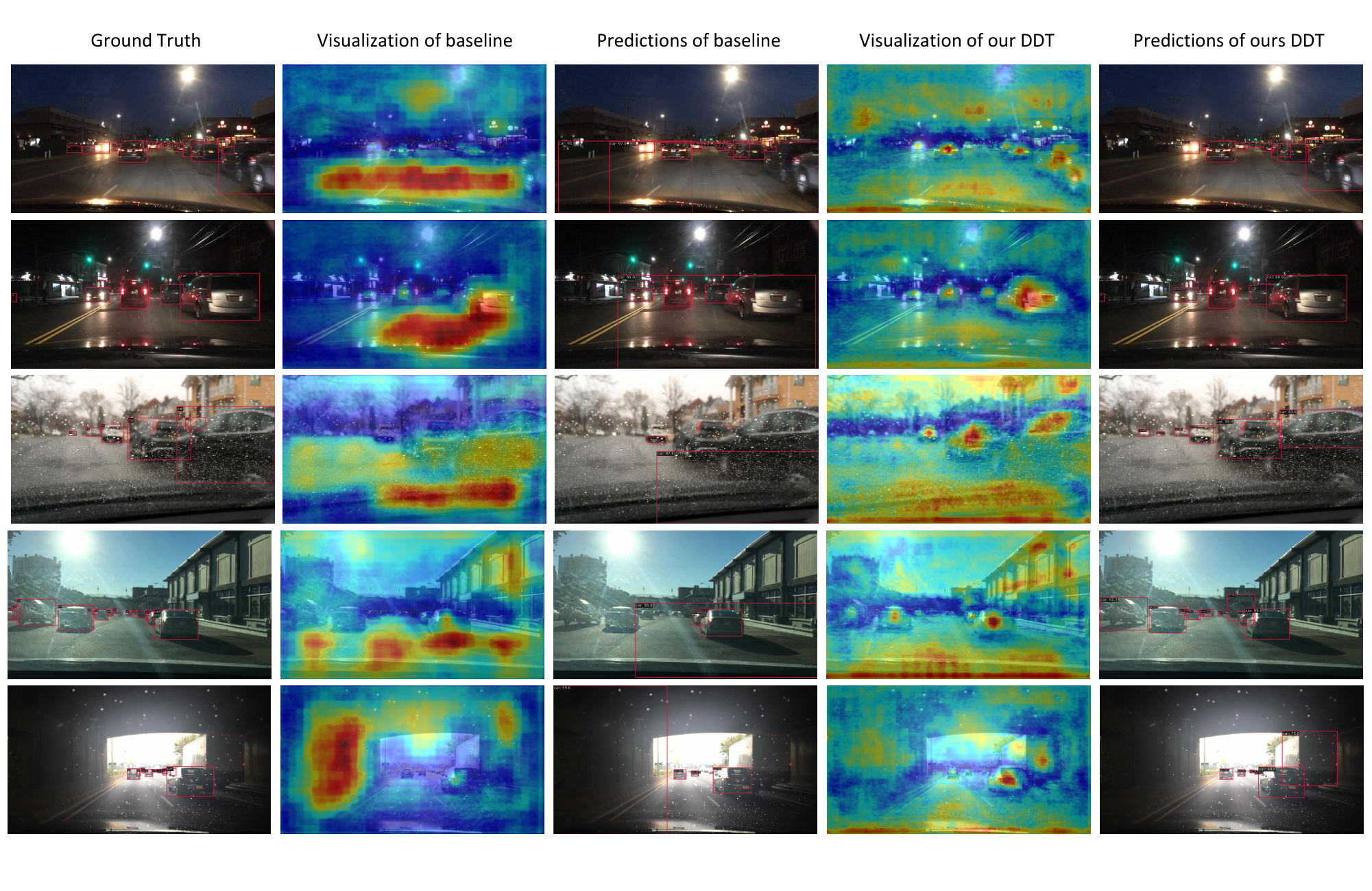}
 \caption{Qualitative prediction results and feature visualization of baseline and our DDT from Sim10K to BDD100K.}
 \label{fig:bdd100k_for_sim10k}
\end{figure*}

\begin{figure*}[t]
  % 调整间距
 \centering
 \includegraphics[width=0.9\linewidth]{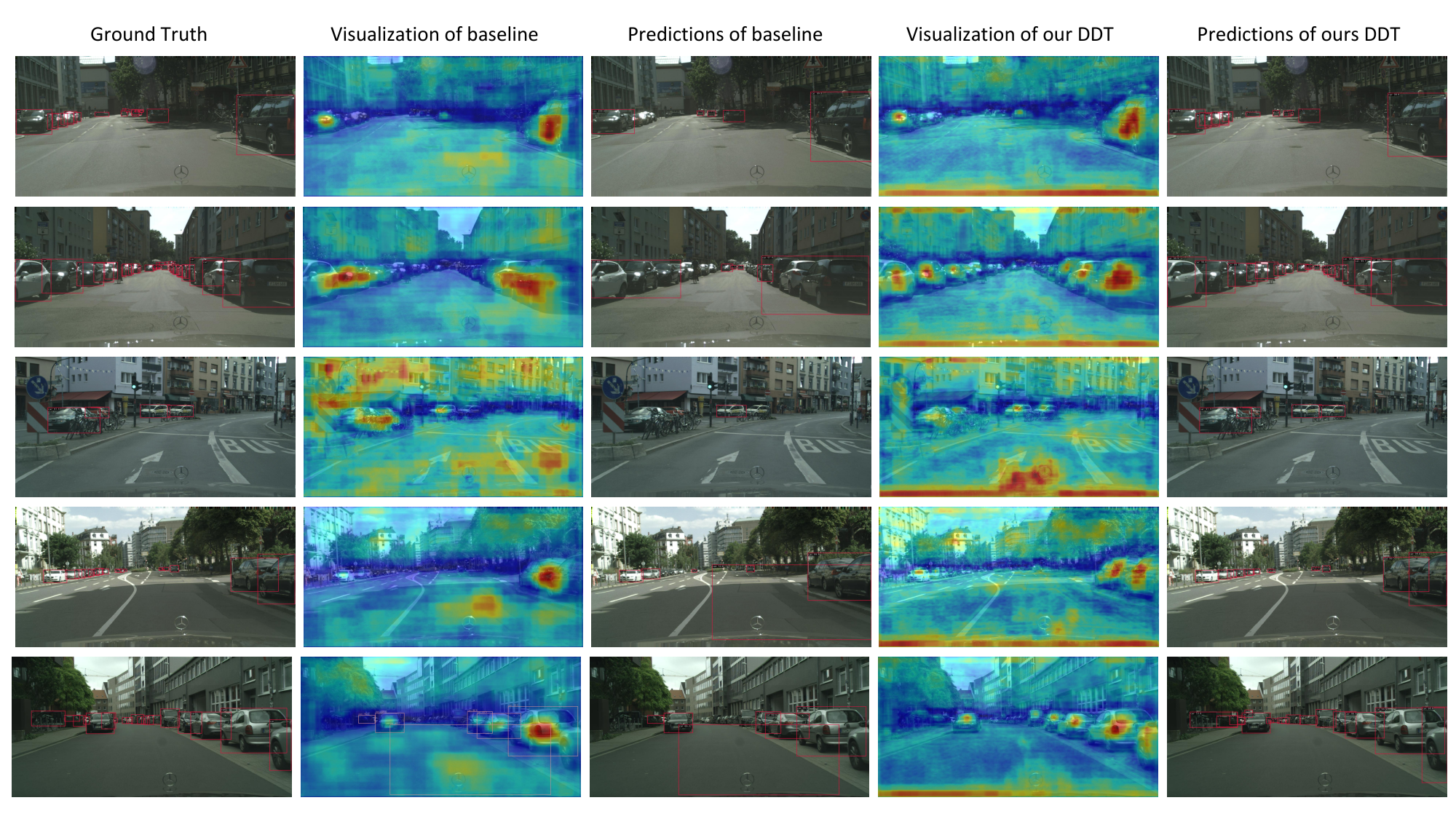}
 \caption{Qualitative prediction results and feature visualization of baseline and our DDT from Sim10K to Cityscapes.}
 \label{fig:city_for_sim10k}
\end{figure*}

\begin{figure*}[t]
  % 调整间距
 \centering
 \includegraphics[width=1.0\linewidth]{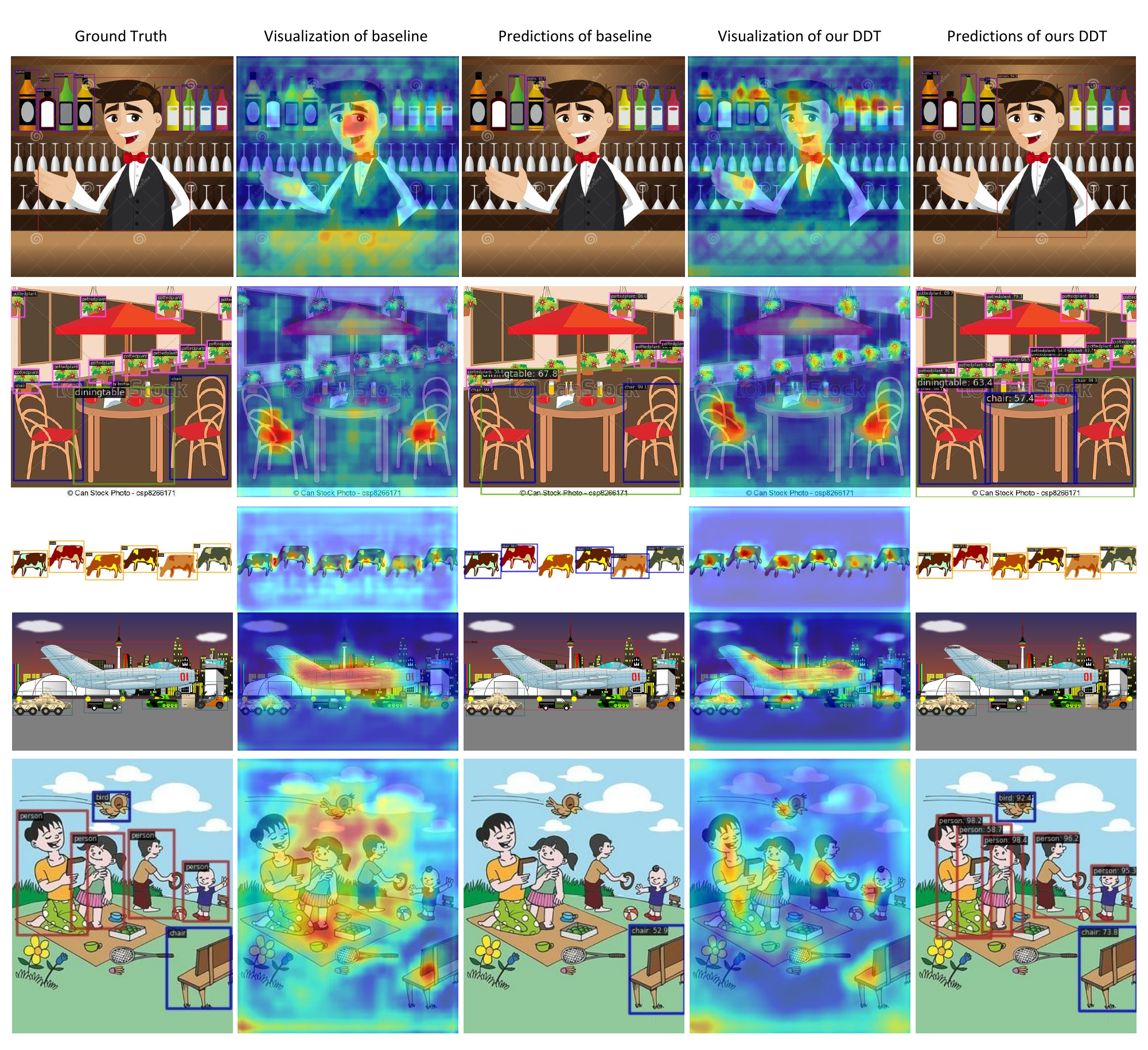}
 \caption{Qualitative prediction results and feature visualization of baseline and our DDT from VOC to Clipart.}
 \label{fig:clipart}
\end{figure*}

\begin{figure*}[t]
  % 调整间距
 \centering
 \includegraphics[width=1.0\linewidth]{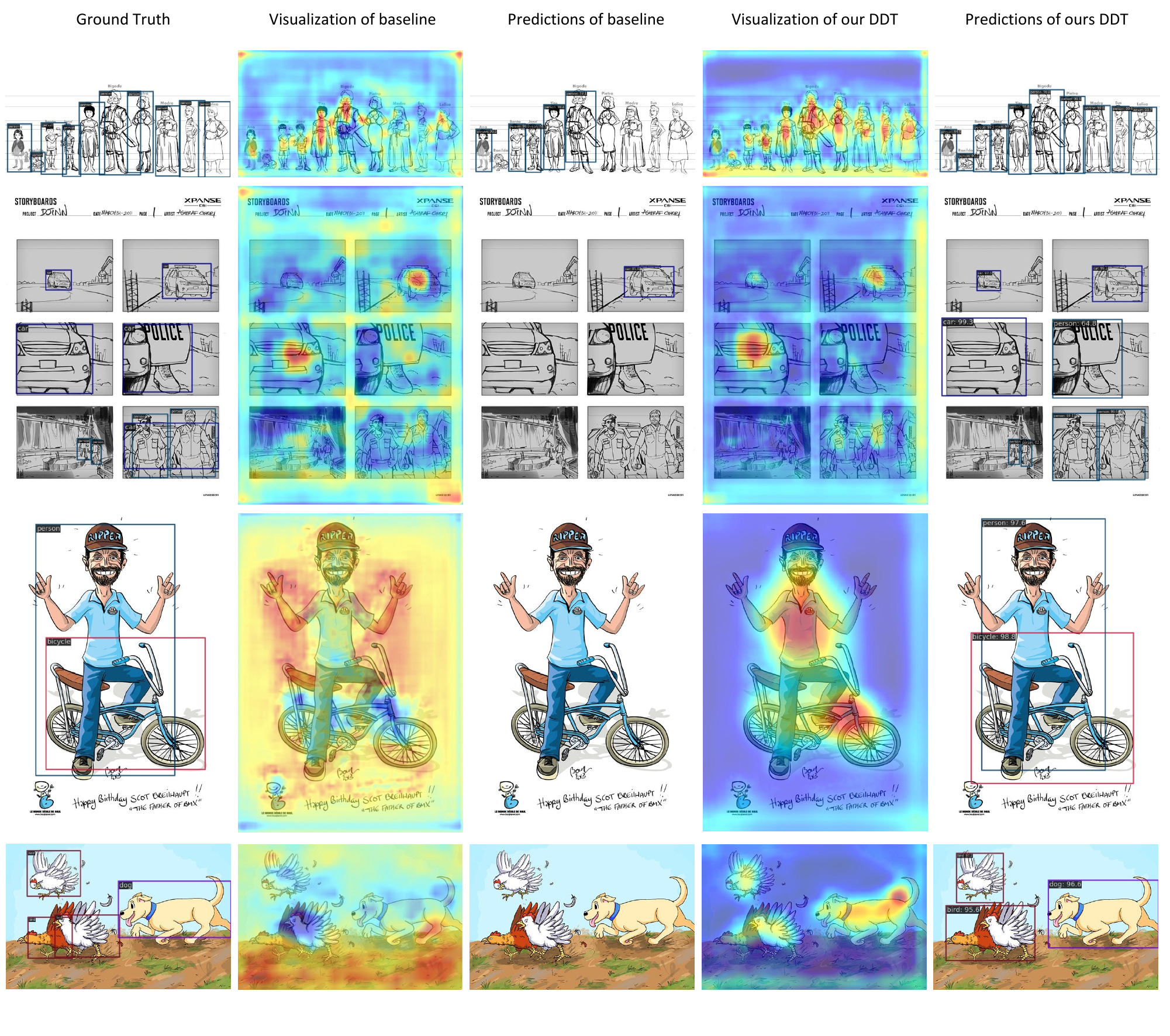}
 \caption{Qualitative prediction results and feature visualization of baseline and our DDT from VOC to Comic.}
 \label{fig:comic}
\end{figure*}

\begin{figure*}[t]
  % 调整间距
 \centering
 \includegraphics[width=1.0\linewidth]{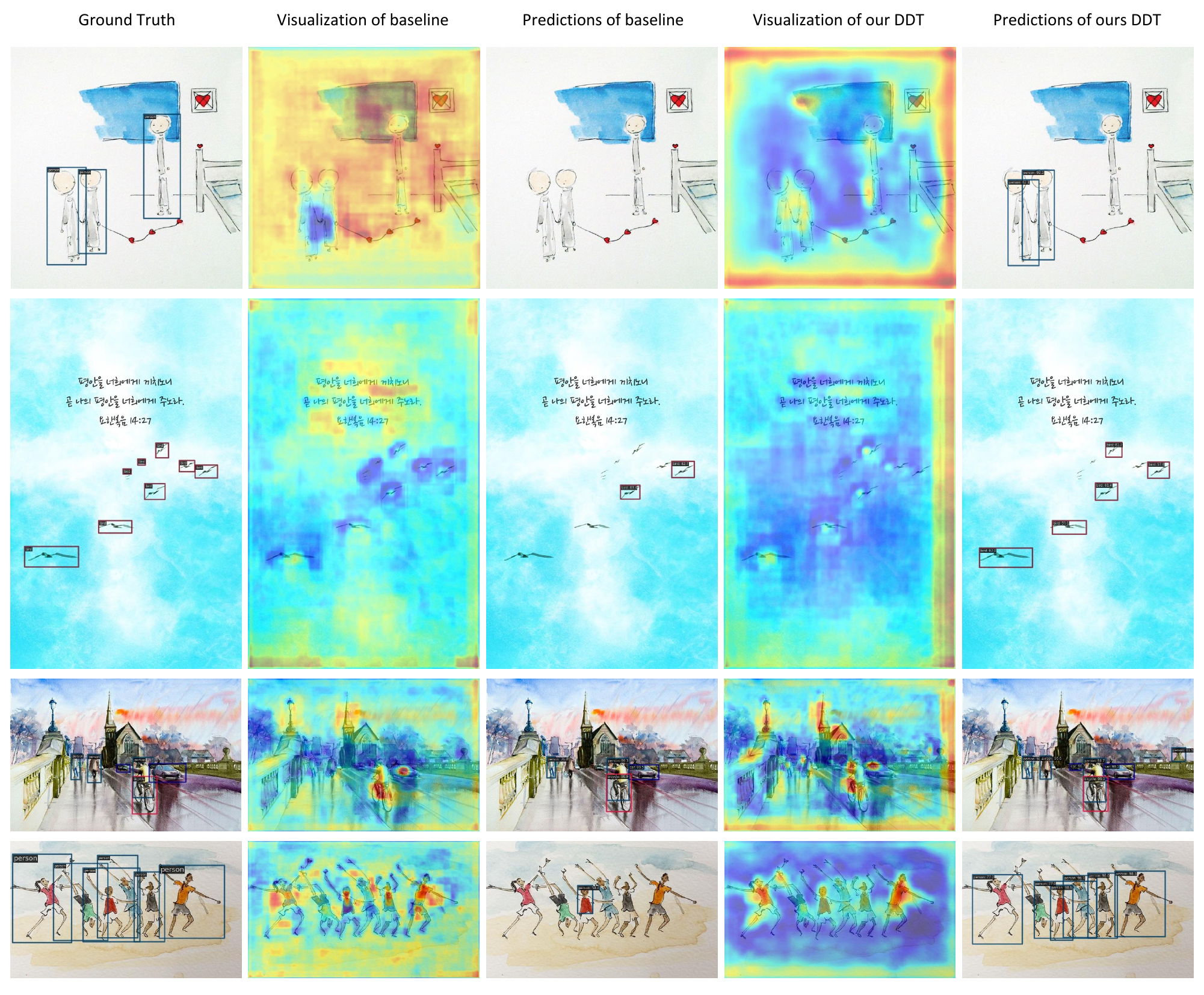}
 \caption{Qualitative prediction results and feature visualization of baseline and our DDT from VOC to Watercolor.}
 \label{fig:watercolor}
\end{figure*}

In Fig. \ref{fig:bdd100k_for_cityscapes}, \ref{fig:bdd100k_for_sim10k}, \ref{fig:city_for_sim10k}, \ref{fig:clipart}, \ref{fig:comic}, \ref{fig:watercolor}, we present additional visualization results for Cityscapes to BDD100K (\textbf{Cs}$\rightarrow$\textbf{B}), Sim10K to BDD100K (\textbf{S}$\rightarrow$\textbf{B}), Sim10K to Cityscapes (\textbf{S}$\rightarrow$\textbf{Cs}),   VOC to Clipart (\textbf{V}$\rightarrow$\textbf{Ca}), VOC to Comic (\textbf{V}$\rightarrow$\textbf{Co}), and VOC to Watercolor (\textbf{V}$\rightarrow$\textbf{W}).

% \usepackage{multirow}
\begin{table*}[]
  \renewcommand{\arraystretch}{1.0}
  \setlength{\tabcolsep}{6pt}
  \centering
  \caption{Quantitative results on adaptation from \textbf{Cityscapes to FoggyCityscapes}. \normalsize The bold indicates the best results.}
  \label{tab:city_to_foggy}
  \begin{tabular}{ccc|ccccccccc}
  \toprule
  Method    & Reference          & Detector                    & bus  & bicycle & car  & mcycle & person & rider & train & truck & mAP  \\ \midrule
  \textbf{UMT} \cite{umt}       & \small{\textit{CVPR'21}}            & FRCNN-\footnotesize R101                  & 56.5 & 37.3    & 48.6 & 30.4   & 33.0   & 46.7  & 46.8  & 34.1  & 41.7 \\
  \textbf{IIOD} \cite{iiod}      & \small{\textit{TPAMI'21}}            & FRCNN-\footnotesize V16                   & 46.1 & 35.3    & 49.6 & 29.9   & 32.8   & 44.4  & 38.0  & 33.0  & 38.6 \\
  \textbf{SADA} \cite{sada}     & \small{\textit{IJCV'21}}             & FRCNN-\footnotesize R50                   & 50.3 & 45.4    & 62.1 & 32.4   & 48.5   & 52.6  & 31.5  & 29.5  & 44.0 \\
  \textbf{CDG} \cite{cdg}      & \small{\textit{AAAI'21}}             & FRCNN-\footnotesize V16                   & 47.5 & 38.9    & 53.1 & 38.3   & 38.0   & 47.4  & 41.1  & 34.2  & 42.3 \\
  \textbf{UaDAN} \cite{uadan}    & \small{\textit{TMM'21}}              & FRCNN-\footnotesize R50                   & 49.4 & 38.9    & 53.6 & 32.3   & 36.5   & 46.1  & 42.7  & 28.9  & 41.1 \\
  \textbf{VDD} \cite{vdd}      & \small{\textit{ICCV'21}}             & FRCNN-\footnotesize V16                   & 52.0 & 36.8    & 51.7 & 34.2   & 33.4   & 44.0  & 34.7  & 33.9  & 40.0 \\
  \textbf{O2Net} \cite{o2net}    & \small{\textit{ACMMM'22}}            & DDETR-\footnotesize R50                   & 47.6 & 45.9    & 63.6 & 38.0   & 48.7   & 51.5  & 47.8  & 31.1  & 46.8 \\
  \textbf{SSAL} \cite{ssal}      & \small{\textit{NeurIPS'22}}          & FCOS                        & 50.0 & 38.7    & 59.4 & 26.0   & 45.1   & 47.4  & 25.7  & 24.5  & 39.6 \\
  \textbf{DDF} \cite{ddf}      & \small{\textit{TMM'22}}              & FRCNN-\footnotesize R50                   & 50.4 & 39.8    & 56.1 & 31.1   & 37.6   & 45.5  & 47.0  & 30.7  & 42.3 \\
  \textbf{D-ADAPT} \cite{dadapt}  & \small{\textit{ICLR'22}}             & FRCNN-\footnotesize R50                   & 36.3 & 46.1    & 61.7 & 37.3   & 44.9   & 54.2  & 24.7  & 25.6  & 42.2 \\
  \textbf{SCAN} \cite{scan}      & \small{\textit{AAAI'22}}             & FCOS-\footnotesize R50                    & 48.6 & 37.3    & 57.3 & 31.0   & 41.7   & 43.9  & 48.7  & 28.7  & 42.1 \\
  \textbf{SIGMA} \cite{li2022sigma}    & \small{\textit{CVPR'22}}             & FCOS-\footnotesize R50                    & 50.4 & 40.6    & 60.3 & 31.7   & 44.0   & 43.9  & 51.5  & 31.6  & 44.2 \\
  \textbf{TIA} \cite{tia}      & \small{\textit{CVPR'22}}             & FRCNN-\footnotesize V16                   & 52.1 & 38.1    & 49.7 & 37.7   & 34.8   & 46.3  & 48.6  & 31.1  & 42.3 \\
  \textbf{TDD} \cite{tdd}      & \small{\textit{CVPR'22}}             & FRCNN-\footnotesize V16                   & 53.0 & 49.1    & 68.2 & 38.9   & 50.7   & 53.7  & 45.1  & 35.1  & 49.2 \\
  \textbf{NLTE} \cite{nlte}      & \small{\textit{CVPR'22}}             & FRCNN-\footnotesize R50                   & 56.7 & 43.3    & 58.7 & 33.7   & 43.1   & 50.7  & 42.7  & 33.6  & 45.4 \\
  \textbf{LODS} \cite{lods}     & \small{\textit{CVPR'22}}             & FRCNN-\footnotesize V16                   & 39.7 & 37.8    & 48.8 & 33.2   & 34.0   & 45.7  & 19.6  & 27.3  & 35.8 \\
  \textbf{PSN} \cite{psn}      & \small{\textit{CVPR'22}}             & FRCNN-\footnotesize V16                   & 48.7 & 39.2    & 53.0 & 33.1   & 37.4   & 45.2  & 38.8  & 31.1  & 40.9 \\
  \textbf{MGA} \cite{zhou2022mga}      & \small{\textit{CVPR'22}}             & FCOS-\footnotesize R101                   & 53.2 & 36.9    & 61.5 & 27.9   & 43.1   & 47.3  & 50.3  & 30.2  & 43.8 \\
  \textbf{MTTrans} \cite{MTTrans}  & \small{\textit{ECCV'22}}             & DDETR-\footnotesize R50                   & 45.9 & 46.5    & 65.2 & 32.6   & 47.7   & 49.9  & 33.8  & 25.8  & 43.4 \\
  \textbf{OADA} \cite{oada}     & \small{\textit{ECCV'22}}             & FCOS-\footnotesize V16                    & 48.5 & 39.8    & 62.9 & 34.3   & 47.8   & 46.5  & 50.9  & 32.1  & 45.4 \\
  \textbf{SCAN++} \cite{scan}   & \small{\textit{TMM'22}}              & FCOS-\footnotesize R101                   & 48.1 & 39.5    & 57.9 & 30.1   & 44.2   & 43.9  & 51.2  & 28.2  & 42.8 \\
  \textbf{MIC}  \cite{mic}     & \small{\textit{CVPR'23}}             & FRCNN-\footnotesize R101                  & 52.4 & 47.5    & 67.0 & 40.6   & 50.9   & 55.3  & 33.7  & 33.9  & 47.6 \\
  \textbf{SIGMA++} \cite{li2023sigma++}   & \small{\textit{TPAMI'23}}            & FRCNN-\footnotesize V16                   & 52.2 & 39.9    & 61.0 & 34.8   & 46.4   & 45.1  & 44.6  & 32.1  & 44.5 \\
  \textbf{CIGAR} \cite{CIGAR}     & \small{\textit{CVPR'23}}             & FCOS-\footnotesize V16                    & 56.6 & 41.3    & 62.1 & 33.7   & 46.1   & 47.3  & 44.3  & 27.8  & 44.9 \\
  \textbf{CMT} \cite{cao2023cmt}      & \small{\textit{CVPR'23}}             & FRCNN-\footnotesize V16                   & 66.0 & 51.2    & 63.7 & 41.4   & 45.9   & 55.7  & 38.8  & 39.6  & 50.3 \\
  \textbf{HT} \cite{deng2023HT}      & \small{\textit{CVPR'23}}             & FCOS-\footnotesize V16                    & 55.9 & 50.3    & 67.5 & 40.1   & 52.1   & 55.8  & 49.1  & 32.7  & \textbf{50.4} \\ \midrule \midrule
  Baseline    & \multirow{2}{*}{/} & \multirow{2}{*}{FRCNN-\footnotesize R18}  & 38.6 & 31.3    & 45.6 & 26.1   & 37.6   & 45.6  & 13.9  & 17.6  & 32.0 \\
  \textbf{DDT(Ours)} &                    &                             & 49.4 & 44.0    & 59.0 & 36.3   & 47.9   & 56.5  & 27.8  & 30.0  & 43.8\footnotesize \textcolor{red}{\textbf{+11.8}} \\
  Baseline    & \multirow{2}{*}{/} & \multirow{2}{*}{FRCNN-\footnotesize R50}  & 39.1 & 32.0    & 42.2 & 23.8   & 36.4   & 44.6  & 14.7  & 19.7  & 31.6 \\
  \textbf{DDT(Ours)} &                    &                             & 53.2 & 51.5    & 63.8 & 44.1   & 50.3   & 59.3  & 41.7  & 33.1  & 49.6\footnotesize \textcolor{red}{\textbf{+18.0}} \\
  Baseline    & \multirow{2}{*}{/} & \multirow{2}{*}{FRCNN-\footnotesize R101} & 35.7 & 31.9    & 41.6 & 23.8   & 34.9   & 41.9  & 5.7   & 19.7  & 29.4 \\
  \textbf{DDT(Ours)} &                    &                             & 53.5 & 52.2    & 64.2 & 43.5   & 50.9   & 60.0  & 42.4  & 33.6  & 50.0\footnotesize \textcolor{red}{\textbf{+21.6}} \\ \bottomrule
  \end{tabular}
  \end{table*}

\begin{table*}[]
  \renewcommand{\arraystretch}{1.15}
  \setlength{\tabcolsep}{6pt}
  \centering
  \caption{Quantitative results on adaptation from \textbf{Cityscapes to BDD100K (Cs$\rightarrow$B)} with FCOS. The bold indicates the best results.}
  \label{tab:city_to_bdd100k_fcos}
    \begin{tabular}{ccc|ccccccc|c}
      \toprule
      Method                                       & Reference                      & Detector                          & bicycle & bus  & car  & mcycle & person & rider & truck & mAP                                                \\ \midrule
      \textbf{DA-Faster} ~\cite{chen2018da-faster} & \small{\textit{CVPR'18}}  & FRCNN\footnotesize -V16                   & 22.4    & 18.0 & 44.2 & 14.2   & 28.9   & 27.4  & 19.1  & 24.9                                               \\
      \textbf{SWDA} ~\cite{saito2019swda}          & \small{\textit{CVPR'19}}  & FRCNN\footnotesize -V16                   & 23.1    & 20.7 & 44.8 & 15.2   & 29.5   & 29.9  & 20.2  & 26.2                                               \\
      \textbf{SCDA} ~\cite{zhu2019scda}            & \small{\textit{CVPR'19}}  & FRCNN\footnotesize -V16                   & 23.2    & 19.6 & 44.4 & 14.8   & 29.3   & 29.2  & 20.3  & 25.8                                               \\
      \textbf{CRDA} ~\cite{crda}                   & \small{\textit{CVPR'20}}  & FRCNN\footnotesize -R101                  & 25.5    & 20.6 & 45.8 & 14.9   & 32.8   & 29.3  & 22.7  & 27.4                                               \\
      \textbf{SED} ~\cite{sed}                     & \small{\textit{AAAI'21}}  & FRCNN\footnotesize -V16                   & 25.0    & 23.4 & 50.4 & 18.9   & 32.4   & 32.6  & 20.6  & 29.0                                               \\
      \textbf{TDD} ~\cite{tdd}                     & \small{\textit{CVPR'22}}  & FRCNN\footnotesize -V16                   & 28.8    & 25.5 & 53.9 & 24.5   & 39.6   & 38.9  & 24.1  & 33.6                                               \\
      \textbf{PT} ~\cite{pt}                       & \small{\textit{ICML'22}}  & FRCNN\footnotesize -V16                   & 28.8    & 33.8 & 52.7 & 23.0   & 40.5   & 39.9  & 25.8  & 34.9                                               \\
      \textbf{EPM} ~\cite{hsu2020epm}              & \small{\textit{ECCV'20}}  & FCOS\footnotesize -R101                   & 20.1    & 19.1 & 55.8 & 14.5   & 39.6   & 26.8  & 18.8  & 27.8                                               \\
      \textbf{SIGMA} ~\cite{li2022sigma}           & \small{\textit{CVPR'22}}  & FCOS\footnotesize -R50                    & 26.3    & 23.6 & 64.1 & 17.9   & 46.9   & 29.6  & 20.2  & 32.7                                               \\
      \textbf{SIGMA++} ~\cite{li2023sigma++}       & \small{\textit{TPAMI'23}} & FRCNN\footnotesize -V16                   & 27.1    & 26.3 & 65.6 & 17.8   & 47.5   & 30.4  & 21.1  & 33.7                                               \\
      \textbf{NSA} ~\cite{nsa}                     & \small{\textit{ICCV'23}}  & FRCNN\footnotesize -V16                   & /       & /    & /    & /      & /      & /     & /     & 35.5                                               \\
      \textbf{HT} ~\cite{deng2023HT}               & \small{\textit{CVPR'23}}  & FCOS\footnotesize -V16                    & 38.0    & 30.6 & 63.5 & 28.2   & 53.4   & 40.4  & 27.4  & 40.2                                               \\ \midrule \midrule
      Baseline    & \multirow{2}{*}{/} & \multirow{2}{*}{FCOS-\footnotesize R18}  & 18.7    & 12.6          & 49.0 & 11.0   & 40.1   & 23.4  & 14.5  & 24.2 \\
      \textbf{DDT(Ours)} &                    &                            & 35.7    & 26.2          & 63.2 & 24.3   & 53.5   & 35.7  & 27.0  & 37.9\footnotesize \textcolor{red}{\textbf{+13.7}} \\
      Baseline    & \multirow{2}{*}{/} & \multirow{2}{*}{FCOS-\footnotesize R50}  & 21.7    & 15.9          & 49.1 & 13.7   & 40.4   & 26.6  & 14.6  & 26.0 \\
      \textbf{DDT(Ours)} &                    &                            & 38.0    & 32.0          & 64.0 & 25.9   & 55.9   & 36.8  & 29.3  & 40.3\footnotesize \textcolor{red}{\textbf{+14.3}} \\
      Baseline    & \multirow{2}{*}{/} & \multirow{2}{*}{FCOS-\footnotesize R101} & 27.0    & 16.4          & 51.4 & 14.7   & 44.0   & 28.8  & 21.2  & 29.1 \\
      \textbf{DDT(Ours)} &                    &                            & 37.9    & 36.1          & 64.5 & 30.8   & 56.9   & 38.7  & 31.8  & \textbf{42.4}\footnotesize \textcolor{red}{\textbf{+13.3}} \\ \bottomrule
    \end{tabular}%
\end{table*}
%%%%%%%%%%%%%%%%%%%%%%%%%%%%%%%%%%%%%%%%%%%%%%%

\begin{table}[]
  % 调整间距
  \renewcommand{\arraystretch}{1.15}
 \setlength{\tabcolsep}{4pt}
 \centering
 \caption{Quantitative results on adaptation from \textbf{Sim10K to BDD100K (S$\rightarrow$B)} with FCOS. \normalsize The bold indicates the best results.}
 \label{tab:sim10k_to_bdd100k_fcos}
 \begin{tabular}{ccc|c}
   \toprule
   Method                              & Reference                     & Detector                          & mAP(car)                                           \\ \midrule
   \textbf{SWDA} ~\cite{saito2019swda} & \small{\textit{CVPR'19}} & FRCNN\footnotesize -V16                   & 42.9                                               \\
   \textbf{CDN} ~\cite{cdn}            & \small{\textit{ECCV'20}} & FRCNN\footnotesize -V16                   & 45.3                                               \\ \midrule \midrule
   Baseline                            & \multirow{2}{*}{/}            & \multirow{2}{*}{FCOS\footnotesize -R18}  & 36.5                                               \\
   \textbf{DDT(Ours)}                  &                               &                                   & 56.0\footnotesize \textcolor{red}{\textbf{+19.5}}          \\
   Baseline                            & \multirow{2}{*}{/}            & \multirow{2}{*}{FCOS\footnotesize -R50}  & 38.7                                             \\
   \textbf{DDT(Ours)}                  &                               &                                   & 56.2\footnotesize \textcolor{red}{\textbf{+17.5}}          \\
   Baseline                            & \multirow{2}{*}{/}            & \multirow{2}{*}{FCOS\footnotesize -R101} & 36.5                                               \\
   \textbf{DDT(Ours)}                  &                               &                                   & \textbf{57.4}\footnotesize \textcolor{red}{\textbf{+20.9}} \\ \bottomrule
 \end{tabular}
\end{table}

\begin{table}[]
% 调整间距
\renewcommand{\arraystretch}{1.15}
\setlength{\tabcolsep}{4pt}
\centering
\caption{Quantitative results on adaptation from Sim10K to Cityscapes (S$\rightarrow$Cs) with FCOS. The bold indicates the best results.}
\label{tab:sim10k_to_city_fcos}
\begin{tabular}{ccc|c}
 \toprule
 Method                                  & Reference                        & Detector                          & mAP(car)                                  \\ \midrule
 \textbf{DA-Faster} ~\cite{chen2018da-faster} & \small{\textit{CVPR'18}}    & FRCNN\footnotesize -V16                   & 39.0                                      \\
 \textbf{SWDA} ~\cite{saito2019swda}          & \small{\textit{CVPR'19}}    & FRCNN\footnotesize -V16                   & 40.7                                      \\
 \textbf{HTCN} ~\cite{htcn}                   & \small{\textit{CVPR'20}}    & FRCNN\footnotesize -R101                  & 42.5                                      \\
 \textbf{UMT} ~\cite{umt}                     & \small{\textit{CVPR'21}}    & FRCNN\footnotesize -R101                  & 43.1                                      \\
 \textbf{SSAL} ~\cite{ssal}              & \small{\textit{NeurIPS'22}} & FCOS\footnotesize -R50                    & 51.8                                      \\
 \textbf{O2NET} ~\cite{o2net}            & \small{\textit{ACMMM'22}}   & DDETR\footnotesize -R50                   & 54.1                                      \\
 \textbf{DDF} ~\cite{ddf}                & \small{\textit{TMM'22}}     & FRCNN\footnotesize -R50                   & 44.3                                      \\
 \textbf{D-ADAPT} ~\cite{dadapt}         & \small{\textit{ICLR'22}}    & FRCNN\footnotesize -R50                   & 51.9                                      \\
 \textbf{SCAN} ~\cite{scan}              & \small{\textit{AAAI'22}}    & FCOS\footnotesize -V16                    & 52.6                                      \\
 \textbf{MTTrans} ~\cite{MTTrans}        & \small{\textit{ECCV'22}}    & DDETR\footnotesize -R50                   & 57.9                                      \\
 \textbf{SIGMA} ~\cite{li2022sigma}      & \small{\textit{CVPR'22}}    & FCOS\footnotesize -R50                    & 53.7                                      \\
 \textbf{TDD}  ~\cite{li2023sigma++}     & \small{\textit{CVPR'22}}    & FRCNN\footnotesize -V16                   & 53.4                                      \\
 \textbf{MGA}  ~\cite{zhou2022mga}       & \small{\textit{CVPR'22}}    & FCOS\footnotesize -R101                   & 54.1                                      \\
 \textbf{OADA}  ~\cite{oada}             & \small{\textit{ECCV'22}}    & FCOS\footnotesize -V16                    & 59.2                                      \\
 \textbf{SIGMA++}  ~\cite{li2023sigma++} & \small{\textit{TPAMI'23}}   & FCOS\footnotesize -V16                    & 53.7                                      \\
 \textbf{CIGAR} ~\cite{CIGAR}            & \small{\textit{CVPR'23}}    & FCOS\footnotesize -V16                    & 58.5                                      \\
 \textbf{NSA} ~\cite{nsa}                & \small{\textit{ICCV'23}}    & FRCNN\footnotesize -V16                   & 56.3                                      \\
 \textbf{HT} ~\cite{deng2023HT}          & \small{\textit{CVPR'23}}    & FRCNN\footnotesize -V16                   & \textbf{65.5}                             \\ \midrule \midrule
 Baseline                                & \multirow{2}{*}{/}               & \multirow{2}{*}{FCOS\footnotesize -R18}  & 47.0                                      \\
 \textbf{DDT(Ours)}                      &                                  &                                   & 61.4\footnotesize \textcolor{red}{\textbf{+14.4}} \\
 Baseline                                & \multirow{2}{*}{/}               & \multirow{2}{*}{FCOS\footnotesize -R50}  & 48.4                                     \\
 \textbf{DDT(Ours)}                      &                                  &                                   & 62.5\footnotesize \textcolor{red}{\textbf{+14.1}} \\
 Baseline                                & \multirow{2}{*}{/}               & \multirow{2}{*}{FCOS\footnotesize -R101} & 51.5                                     \\
 \textbf{DDT(Ours)}                      &                                  &                                   & 63.5\footnotesize \textcolor{red}{\textbf{+12.0}} \\ \bottomrule
\end{tabular}%
\end{table}

\begin{table}[]
\renewcommand{\arraystretch}{1.15}
\setlength{\tabcolsep}{2pt}
 % 调整间距
\centering
\caption{Quantitative results on adaptation from VOC to Comic (V$\rightarrow$Co) with FCOS. The bold indicates the best results.}

\label{tab:voc_to_comic_fcos}
\resizebox{\columnwidth}{!}{%
  \begin{tabular}{ccc|cccccc|c}
    \toprule
    Method                                       & Reference                     & Detector                          & bicycle & bird & car  & cat  & dog  & person & mAP                                                \\ \midrule
    \textbf{DA-Faster} ~\cite{chen2018da-faster} & \small{\textit{CVPR'18}} & FRCNN\footnotesize -V16                   & 31.1    & 10.3 & 15.5 & 12.4 & 19.3 & 39.0   & 21.2                                               \\
    \textbf{SWDA} ~\cite{saito2019swda}          & \small{\textit{CVPR'19}} & FRCNN\footnotesize -V16                   & 36.4    & 21.8 & 29.8 & 15.1 & 23.5 & 49.6   & 29.4                                               \\
    \textbf{STABR} ~\cite{stabr}                 & \small{\textit{CVPR'19}} & SSD\footnotesize -V16                     & 50.6    & 13.6 & 31.0 & 7.5  & 16.4 & 41.4   & 26.8                                               \\
    \textbf{MCRA}   ~\cite{mcda}                 & \small{\textit{ECCV'20}} & FRCNN\footnotesize -V16                   & 47.9    & 20.5 & 37.4 & 20.6 & 24.5 & 50.2   & 33.5                                               \\
    \textbf{I3Net} ~\cite{i3net}                 & \small{\textit{CVPR'21}} & SSD\footnotesize -V16                     & 47.5    & 19.9 & 33.2 & 11.4 & 19.4 & 49.1   & 30.1                                               \\
    \textbf{DBGL}  ~\cite{dbgl}                  & \small{\textit{ICCV'21}} & FRCNN\footnotesize -R101                  & 35.6    & 20.3 & 33.9 & 16.4 & 26.6 & 45.3   & 29.7                                               \\
    \textbf{D-ADAPT} ~\cite{dadapt}              & \small{\textit{ICLR'22}} & FRCNN\footnotesize -R101                  & 52.4    & 25.4 & 42.3 & 43.7 & 25.7 & 53.5   & 40.5                                               \\ \midrule \midrule
    Baseline  & \multirow{2}{*}{/} & \multirow{2}{*}{FCOS-\footnotesize R18}  & 17.7    & 7.1  & 8.3  & 2.4  & 5.7  & 25.3   & 11.1 \\
    \textbf{DDT(Ours)} &                    &                            & 54.2    & 26.6 & 45.5 & 29.8 & 34.4 & 72.2   & 43.8\footnotesize \textcolor{red}{\textbf{+32.7}} \\
    Baseline  & \multirow{2}{*}{/} & \multirow{2}{*}{FCOS-\footnotesize R50}  & 20.9    & 7.2  & 11.3 & 4.7  & 7.9  & 27.0   & 13.2 \\
    \textbf{DDT(Ours)} &                    &                            & 55.1    & 32.2 & 51.4 & 33.8 & 38.7 & 74.6   & 47.6\footnotesize \textcolor{red}{\textbf{+34.4}} \\
    Baseline  & \multirow{2}{*}{/} & \multirow{2}{*}{FCOS-\footnotesize R101} & 26.0    & 9.5  & 15.4 & 7.3  & 8.0  & 29.3   & 15.9 \\
    \textbf{DDT(Ours)} &                    &                            & 55.6    & 38.2 & 55.6 & 36.6 & 48.1 & 75.9   & \textbf{51.6}\footnotesize \textcolor{red}{\textbf{+35.7}} \\ \bottomrule
  \end{tabular}%
}
\end{table}

\begin{table}[]
\renewcommand{\arraystretch}{1.15}
\setlength{\tabcolsep}{1pt}
 % 调整间距
\centering
\caption{Quantitative results on adaptation from VOC to Watercolor (V$\rightarrow$W) with FCOS. The bold indicates the best results.}

\label{tab:voc_to_watercolor_fcos}
\resizebox{\columnwidth}{!}{%
  \begin{tabular}{ccc|cccccc|c}
    \toprule
    Method                                   & Reference                      & Detector                          & bicycle & bird & car  & cat  & dog  & person & mAP                                                \\ \midrule
    \textbf{SWDA}  ~\cite{chen2018da-faster} & \small{\textit{CVPR‘19}}  & FRCNN\footnotesize -V16                   & 82.3    & 55.9 & 46.5 & 32.7 & 35.5 & 66.7   & 53.3                                               \\
    \textbf{MCRA} ~\cite{mcra}               & \small{\textit{ECCV‘20}}  & FRCNN\footnotesize -V16                   & 87.9    & 52.1 & 51.8 & 41.6 & 33.8 & 68.8   & 56.0                                               \\
    \textbf{UMT}  ~\cite{umt}                & \small{\textit{CVPR’21}}  & FRCNN\footnotesize -R101                  & 88.2    & 55.3 & 51.7 & 39.8 & 43.6 & 69.9   & 58.1                                               \\
    \textbf{IIOD}  ~\cite{iiod}              & \small{\textit{TPAMI‘21}} & FRCNN\footnotesize -V16                   & 95.8    & 54.3 & 48.3 & 42.4 & 35.1 & 65.8   & 56.9                                               \\
    \textbf{I3Net}  ~\cite{i3net}            & \small{\textit{CVPR’21}}  & SSD\footnotesize -V16                     & 81.1    & 49.3 & 46.2 & 35.0 & 31.9 & 65.7   & 51.5                                               \\
    \textbf{SADA}  ~\cite{sada}              & \small{\textit{IJCV‘21}}  & FRCNN\footnotesize -R50                   & 82.9    & 54.6 & 52.3 & 40.5 & 37.7 & 68.2   & 56.0                                               \\
    \textbf{CDG}  ~\cite{cdg}                & \small{\textit{AAAI’21}}  & FRCNN\footnotesize -V16                   & 97.7    & 53.1 & 52.1 & 47.3 & 38.7 & 68.9   & 59.7                                               \\
    \textbf{VDD}   ~\cite{vdd}               & \small{\textit{ICCV‘21}}  & FRCNN\footnotesize -V16                   & 90.0    & 56.6 & 49.2 & 39.5 & 38.8 & 65.3   & 56.6                                               \\
    \textbf{DBGL}   ~\cite{dbgl}             & \small{\textit{ICCV’21}}  & FRCNN\footnotesize -R101                  & 83.1    & 49.3 & 50.6 & 39.8 & 38.7 & 61.3   & 53.8                                               \\
    \textbf{AT}  ~\cite{li2022AT}            & \small{\textit{CVPR’22}}  & FRCNN\footnotesize -V16                   & 93.6    & 56.1 & 58.9 & 37.3 & 39.6 & 73.8   & 59.9                                               \\
    \textbf{LODS}  ~\cite{lods}              & \small{\textit{CVPR’22}}  & FRCNN\footnotesize -R101                  & 95.2    & 53.1 & 46.9 & 37.2 & 47.6 & 69.3   & 58.2                                               \\ \midrule \midrule
    Baseline  & \multirow{2}{*}{/}       & \multirow{2}{*}{FCOS-\footnotesize R18}  & 69.8    & 34.9 & 37.8 & 23.3 & 16.0 & 48.7   & 38.4 \\
    \textbf{DDT(Ours)} &                          &                            & 80.3    & 60.1 & 52.5 & 42.4 & 34.3 & 75.8   & 57.6\footnotesize \textcolor{red}{\textbf{+19.2}} \\
    Baseline  & \multirow{2}{*}{/}       & \multirow{2}{*}{FCOS-\footnotesize R50}  & 66.9    & 42.4 & 44.6 & 21.5 & 13.7 & 48.3   & 39.6 \\
    \textbf{DDT(Ours)} &                          &                            & 94.4    & 63.1 & 51.8 & 40.8 & 34.3 & 75.9   & 60.1\footnotesize \textcolor{red}{\textbf{+20.5}} \\
    Baseline  & \multirow{2}{*}{/}       & \multirow{2}{*}{FCOS-\footnotesize R101} & 64.0    & 44.3 & 41.8 & 25.7 & 21.5 & 53.6   & 41.8 \\
    \textbf{DDT(Ours)} &                          &                            & 96.9    & 65.6 & 55.4 & 49.6 & 40.5 & 77.2   & \textbf{64.2}\footnotesize \textcolor{red}{\textbf{+22.4}} \\ \bottomrule
  \end{tabular}%
}
\end{table}

\begin{table*}[]
  \renewcommand{\arraystretch}{1.15}
   % 调整间距
  \setlength{\tabcolsep}{1pt}
  \centering
  \caption{Quantitative results on adaptation from VOC to Clipart (V$\rightarrow$Ca) with FCOS. The bold indicates the best results.}
  \label{tab:voc_to_clipart_fcos}
  \resizebox{\textwidth}{!}{%
    \begin{tabular}{ccc|cccccccccccccccccccc|c}
      \toprule
      Method                           & Reference                     & Detector                          & aero & bcycle & bird & boat & bottle & bus  & car  & cat  & chair & cow  & table & dog  & horse & bike & psn  & plant & sheep & sofa & train & tv   & mAP                                       \\ \midrule
      \textbf{SWDA} ~\cite{saito2019swda} & \small{\textit{CVPR'19}}  & FRCNN\footnotesize -V16                   & 26.2 & 48.5   & 32.6 & 33.7 & 38.5   & 54.3 & 37.1 & 18.6 & 34.8  & 58.3 & 17.0  & 12.5 & 33.8  & 65.5 & 61.6 & 52.0  & 9.3   & 24.9 & 54.1  & 49.1 & 38.1                                      \\
      \textbf{CRDA} ~\cite{crda}          & \small{\textit{CVPR'20}}  & FRCNN\footnotesize -R101                  & 28.7 & 55.3   & 31.8 & 26.0 & 40.1   & 63.6 & 36.6 & 9.4  & 38.7  & 49.3 & 17.6  & 14.1 & 33.3  & 74.3 & 61.3 & 46.3  & 22.3  & 24.3 & 49.1  & 44.3 & 38.3                                      \\
      \textbf{HTCN}  ~\cite{htcn}         & \small{\textit{CVPR'20}}  & FRCNN\footnotesize -R101                  & 33.6 & 58.9   & 34.0 & 23.4 & 45.6   & 57.0 & 39.8 & 12.0 & 39.7  & 51.3 & 21.1  & 20.1 & 39.1  & 72.8 & 63.0 & 43.1  & 19.3  & 30.1 & 50.2  & 51.8 & 40.3                                      \\
      \textbf{SAPNet} ~\cite{sapnet}      & \small{\textit{ECCV'20}}  & FRCNN\footnotesize -R101                  & 27.4 & 70.8   & 32.0 & 27.9 & 42.4   & 63.5 & 47.5 & 14.3 & 48.2  & 46.1 & 31.8  & 17.9 & 43.8  & 68.0 & 68.1 & 49.0  & 18.7  & 20.4 & 55.8  & 51.3 & 42.2                                      \\
      \textbf{UMT} ~\cite{umt}            & \small{\textit{CVPR'21}}  & FRCNN\footnotesize -R101                  & 39.6 & 59.1   & 32.4 & 35.0 & 45.1   & 61.9 & 48.4 & 7.5  & 46.0  & 67.6 & 21.4  & 29.5 & 48.2  & 75.9 & 70.5 & 56.7  & 25.9  & 28.9 & 39.4  & 43.6 & 44.1                                      \\
      \textbf{IIOD} ~\cite{iiod}          & \small{\textit{TPAMI'21}} & FRCNN\footnotesize -V16                   & 41.5 & 52.7   & 34.5 & 28.1 & 43.7   & 58.5 & 41.8 & 15.3 & 40.1  & 54.4 & 26.7  & 28.5 & 37.7  & 75.4 & 63.7 & 48.7  & 16.5  & 30.8 & 54.5  & 48.7 & 42.1                                      \\
      \textbf{SADA}  ~\cite{sada}         & \small{\textit{IJCV'21}}  & FRCNN\footnotesize -R50                   & 29.4 & 56.8   & 30.6 & 34.0 & 49.5   & 50.5 & 47.7 & 18.7 & 48.5  & 64.4 & 20.3  & 29.0 & 42.3  & 84.1 & 73.4 & 37.4  & 20.5  & 39.8 & 41.2  & 48.0 & 43.3                                      \\
      \textbf{UaDAN} ~\cite{uadan}        & \small{\textit{TMM'21}}   & FRCNN\footnotesize -R50                   & 35.0 & 72.7   & 41.0 & 24.4 & 21.3   & 69.8 & 53.5 & 2.3  & 34.2  & 61.2 & 31.0  & 29.5 & 47.9  & 63.6 & 62.2 & 61.3  & 13.9  & 7.6  & 48.6  & 23.9 & 40.2                                      \\
      \textbf{DBGL} ~\cite{dbgl}          & \small{\textit{ICCV'21}}  & FRCNN\footnotesize -R50                   & 28.5 & 52.3   & 34.3 & 32.8 & 38.6   & 66.4 & 38.2 & 25.3 & 39.9  & 47.4 & 23.9  & 17.9 & 38.9  & 78.3 & 61.2 & 51.7  & 26.2  & 28.9 & 56.8  & 44.5 & 41.6                                      \\
      \textbf{AT} ~\cite{li2022AT}     & \small{\textit{CVPR'22}} & FRCNN\footnotesize -V16                   & 33.8 & 60.9   & 38.6 & 49.4 & 52.4   & 53.9 & 56.7 & 7.5  & 52.8  & 63.5 & 34.0  & 25.0 & 62.2  & 72.1 & 77.2 & 57.7  & 27.2  & 52.0 & 55.7  & 54.1 & 49.3                                      \\
      \textbf{D-ADAPT} ~\cite{dadapt}  & \small{\textit{ICLR'22}} & FRCNN\footnotesize -R50                   & 56.4 & 63.2   & 42.3 & 40.9 & 45.3   & 77.0 & 48.7 & 25.4 & 44.3  & 58.4 & 31.4  & 24.5 & 47.1  & 75.3 & 69.3 & 43.5  & 27.9  & 34.1 & 60.7  & 64.0 & 49.0                                      \\
      \textbf{TIA} ~\cite{tia}         & \small{\textit{CVPR'22}} & FRCNN\footnotesize -R101                  & 42.2 & 66.0   & 36.9 & 37.3 & 43.7   & 71.8 & 49.7 & 18.2 & 44.9  & 58.9 & 18.2  & 29.1 & 40.7  & 87.8 & 67.4 & 49.7  & 27.4  & 27.8 & 57.1  & 50.6 & 46.3                                      \\
      \textbf{LODS} ~\cite{lods}       & \small{\textit{CVPR'22}} & FRCNN\footnotesize -R101                  & 43.1 & 61.4   & 40.1 & 36.8 & 48.2   & 45.8 & 48.3 & 20.4 & 44.8  & 53.3 & 32.5  & 26.1 & 40.6  & 86.3 & 68.5 & 48.9  & 25.4  & 33.2 & 44.0  & 56.5 & 45.2                                      \\
      \textbf{CIGAR} ~\cite{CIGAR}     & \small{\textit{CVPR'23}} & FCOS\footnotesize -R101                   & 35.2 & 55.0   & 39.2 & 30.7 & 60.1   & 58.1 & 46.9 & 31.8 & 47.0  & 61.0 & 21.8  & 26.7 & 44.6  & 52.4 & 68.5 & 54.4  & 31.3  & 38.8 & 56.5  & 63.5 & 46.2                                      \\
      \textbf{CMT}  ~\cite{cao2023cmt} & \small{\textit{CVPR'23}} & FRCNN\footnotesize -V16                   & 39.8 & 56.3   & 38.7 & 39.7 & 60.4   & 35.0 & 56.0 & 7.1  & 60.1  & 60.4 & 35.8  & 28.1 & 67.8  & 84.5 & 80.1 & 55.5  & 20.3  & 32.8 & 42.3  & 38.2 & 47.0                                      \\ \midrule \midrule
      Baseline  & \multirow{2}{*}{/}       & \multirow{2}{*}{FCOS-\footnotesize R18}   & 18.7 & 26.0   & 15.0 & 10.1 & 19.5   & 65.6 & 30.6 & 1.8  & 24.3  & 4.2  & 24.1  & 7.9  & 24.9  & 42.1 & 33.5 & 26.1  & 0.2   & 17.2 & 23.0  & 11.2 & 21.3 \\
     \textbf{DDT(Ours)} &                          &                             & 48.9 & 58.9   & 32.3 & 30.0 & 42.4   & 72.4 & 54.5 & 11.2 & 48.6  & 38.9 & 30.2  & 27.5 & 40.1  & 87.7 & 76.0 & 53.2  & 33.5  & 38.8 & 49.5  & 47.1 & 46.1\footnotesize \textcolor{red}{\textbf{+24.8}} \\
      Baseline  & \multirow{2}{*}{/}       & \multirow{2}{*}{FCOS-\footnotesize R50}   & 40.0 & 26.7   & 17.8 & 21.0 & 31.9   & 32.2 & 28.8 & 12.2 & 36.3  & 35.7 & 28.3  & 6.1  & 25.5  & 43.1 & 37.2 & 33.5  & 5.1   & 25.6 & 24.5  & 26.3 & 26.9 \\
      \textbf{DDT(Ours)} &                          &                             & 50.7 & 53.1   & 34.1 & 41.5 & 57.0   & 86.3 & 57.1 & 9.3  & 49.5  & 52.8 & 33.6  & 32.4 & 49.0  & 93.1 & 82.1 & 57.8  & 37.1  & 42.6 & 54.1  & 60.8 & 51.7\footnotesize \textcolor{red}{\textbf{+24.8}} \\
      Baseline  & \multirow{2}{*}{/}       & \multirow{2}{*}{FCOS-\footnotesize R101}  & 33.6 & 42.3   & 21.2 & 20.1 & 32.9   & 62.0 & 30.0 & 14.5 & 41.1  & 17.9 & 33.0  & 9.1  & 30.4  & 46.5 & 39.1 & 37.4  & 8.8   & 22.6 & 27.3  & 16.1 & 29.3 \\
      \textbf{DDT(Ours)} &                          &                             & 58.6 & 73.2   & 42.0 & 48.0 & 54.7   & 84.7 & 65.2 & 17.0 & 55.9  & 49.4 & 35.5  & 40.5 & 58.6  & 84.8 & 82.9 & 58.0  & 39.1  & 41.7 & 54.7  & 61.3 & \textbf{55.3}\footnotesize \textcolor{red}{\textbf{+26.0}} \\ \bottomrule
    \end{tabular}%
  }
\end{table*}

\clearpage

\end{document}